\documentclass[letterpaper, 10 pt, journal, twoside]{IEEEtran}

\bstctlcite{bstctl:etal, bstctl:nodash, bstctl:simpurl}

\usepackage{times}
\usepackage{graphicx}
\usepackage{amsmath}
\usepackage{amssymb}
\usepackage{textcomp}
\usepackage{multirow}
\usepackage{booktabs}
\usepackage{flushend}
\usepackage{xcolor}
\usepackage{multicol}
\usepackage{color}
\usepackage{makecell}
\usepackage{bbding}
\usepackage{array}
\usepackage{pifont}
\usepackage{colortbl}
\newcommand{\new}[1]{\textcolor{black}{#1}}

\newcommand{\xmark}{\ding{55}}%
\usepackage[figureposition=bottom,tableposition=top]{caption}
\usepackage[colorlinks,pagebackref=false,citecolor=blue,bookmarks=false,hypertexnames=true]{hyperref}

\ifx01 
\newcommand{\sy}[1]{\textcolor{red}{[SY: #1]}}
\newcommand{\AD}[1]{\textcolor{red}{[Arindam: #1]}}
\newcommand{\brk}[1]{\textcolor{red}{[BRK: #1]}}
\else 
\newcommand{\sy}[1]{\textcolor{red}{}}
\newcommand{\AD}[1]{\textcolor{red}{}}
\newcommand{\brk}[1]{\textcolor{red}{}}
\fi

\begin{document}

\bstctlcite{IEEEexample:BSTcontrol}

\title{
Spatio-Contextual Deep Network Based Multimodal Pedestrian Detection For Autonomous Driving
}

\author{
Kinjal Dasgupta$^{1}$,
Arindam Das$^{2}$, 
Sudip Das$^{1}$, 
Ujjwal Bhattacharya$^{1}$
 and 
Senthil Yogamani$^{3}$ 

\thanks{$^{1}$CVPR Unit, Indian Statistical Institute, Kolkata, India.}
\thanks{$^{2}$Detection Vision Systems, Valeo India.}
\thanks{$^{3}$Valeo Visions Systems, Ireland.}
\thanks{Corresponding Email: arindam.das@valeo.com}
}

\markboth{IEEE Transactions on Intelligent Transportation Systems}
{Dasgupta \MakeLowercase{\textit{et al.}}: Spatio-Contextual Deep Network Based Multimodal Pedestrian Detection For Autonomous Driving}

\maketitle

\begin{abstract}
Pedestrian Detection is the most critical module of an Autonomous Driving system. Although a camera is commonly used for this purpose, its quality degrades severely in low-light night time driving scenarios. On the other hand, the quality of a thermal camera image remains unaffected in similar conditions. This paper proposes an end-to-end multimodal fusion model for pedestrian detection using RGB and thermal images.
Its novel spatio-contextual deep network architecture is capable of exploiting the multimodal input efficiently. It consists of two distinct deformable ResNeXt-50 encoders for feature extraction from the two modalities. Fusion of these two encoded features takes place inside a multimodal feature embedding module (MuFEm) consisting of several groups of a pair of Graph Attention Network and a feature fusion unit.  
The output of the last feature fusion unit of MuFEm is subsequently passed to two CRFs for their spatial refinement. Further enhancement of the features is achieved by applying channel-wise attention and extraction of contextual information with the help of four RNNs traversing in four different directions. 
Finally, these feature maps are used by a single-stage decoder to generate the bounding box of each pedestrian and the score map. We have performed extensive experiments of the proposed framework on three publicly available multimodal pedestrian detection benchmark datasets, namely KAIST, CVC-14, and UTokyo. The results on each of them improved the respective state-of-the-art performance. A short video giving an overview of this work along with its qualitative results can be seen at \url{https://youtu.be/FDJdSifuuCs}. \new{Our source code will be released upon publication of the paper}.

\end{abstract}

\begin{IEEEkeywords}
Pedestrian Detection, Multi-spectral Imagery, Thermal Imagery,  Multimodal Feature Fusion,  Deep Learning.
\end{IEEEkeywords}
\IEEEpeerreviewmaketitle

\section{Introduction}







Object detection is an extensively studied area of Computer Vision, while pedestrian detection is a special case of the same. There are several important applications of automatic pedestrian detection, and the most significant one is autonomous driving. Cameras are the primary sensor as the transportation infrastructure is built for human optical sensing and also because of the dense semantic and geometric information it captures at low cost \cite{horgan2015vision, eising2021near}. Due to the safety requirements of an autonomous driving system, multiple sensors, including camera, lidar, radar, ultrasonic, etc., are commonly used to improve reliability and safety. 
There is extensive research performed on pedestrian detection  \cite{kuutti2020survey}. The majority of pedestrian detection studies \cite{geronimo2010survey} considered images captured by RGB cameras under daylight conditions. 
One of the main challenges is its performance in adverse weather and poor lighting conditions. In particular, night time scenarios with illuminance levels less than 1 lux have poor performance.  The quality of semantic representation of the object can degrade severely with insufficient lighting \cite{dhananjaya2021weather}.
A large proportion of pedestrian-related accidents occurs in the night time in the case of human driving \cite{mang2020}. Thus it is necessary to achieve robust pedestrian detection at night time in an autonomous driving system. This is achieved by making use of multiple complementary sensors and using fusion algorithms to improve the reliability \cite{chavez2015multiple, feng2020deep}.

\begin{figure}
    \centering
    \includegraphics[width=0.5\textwidth]{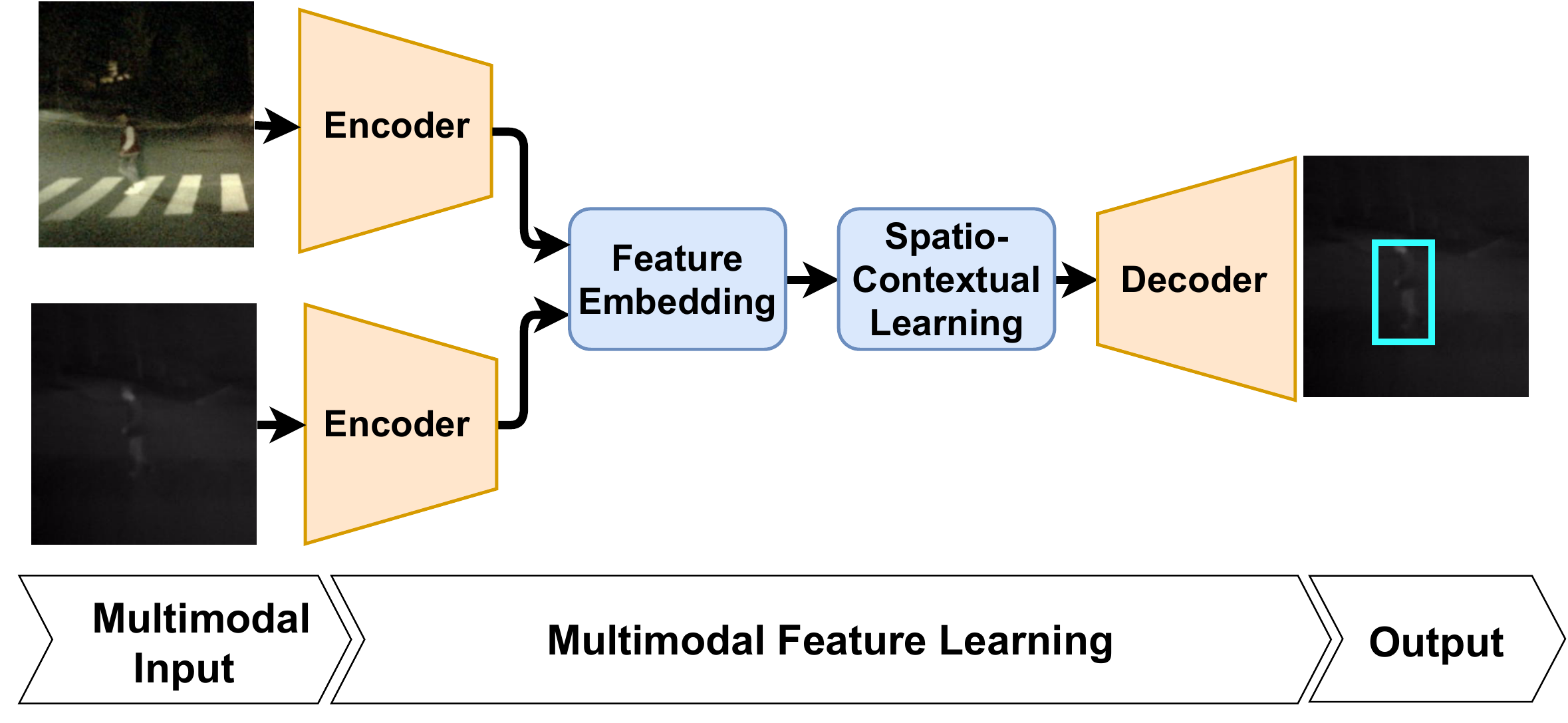}\\
    \caption{Our \textbf{multimodal pedestrian detection model} consists of two unimodal encoders and a single-stage detection decoder. We design two novel modules for efficient feature embedding and spatio-contextual feature representation learning.}

    \vspace{2mm}    
    
    \label{fig:high_level_PD}
\end{figure}

The infrared spectrum lies below the visible spectrum used by RGB cameras. It is categorized into near and far-infrared regions with a region in the middle, further sub-divided into short, mid, and long wavelengths. Mid and long-wavelength infrared are commonly known as thermal infrared as the intensity of captured radiation are correlated with the temperature of an object \cite{miethig2019leveraging}. They are typically passive sensors compared to near-infrared sensors, which require an active source to illuminate the scene. Passive sensors are cost-effective and beginning to be a crucial sensor for autonomous driving. They are commonly used in automotive night vision systems providing a display on the dashboard monitor. Toyota was one of the first to deploy it in 2002 in the Land-Cruiser vehicle. Audi A8 and BMW 7 series released thermal camera-based night vision systems with animal detection feature \cite{Auto-night-vision}. More recent systems support pedestrian detection and animal detection as they have unique thermal signatures relative to other objects. Thermal sensors are also robust in bright sun glare scenarios, which is another major challenge in automated driving \cite{sunglare}.

In this paper, we propose an end-to-end CNN based fusion model for pedestrian detection using both thermal and RGB images as illustrated in Figure \ref{fig:high_level_PD}. Thermal cameras are less variant to ambient light and capture less texture compared to RGB cameras. Thus there is a significant domain gap between them, and an effective fusion strategy is necessary. We use a standard encoder to perform unimodal feature extraction independently for the two input modalities. We propose a multimodal feature embedding module using graph attention network and spatio-contextual feature aggregation module using channel attention and multi-directional RNN to obtain multimodal features. Finally, the multimodal features are input to a single stage detection decoder which produces pedestrian bounding boxes and score maps.

The main contributions of our work include the: 
\begin{enumerate}
    \item Design of a novel multimodal feature embedding module using graph attention network and feature fusion unit to address the modality imbalance problem.
    \item Design of a spatio-contextual feature aggregation module to improve fusion using CRF-based refinement, channel-wise attention, and $4$Dir-IRNN components.
    \item Implementation of an end-to-end trainable network achieving state-of-the-art results on three public datasets, namely KAIST, CVC-$14$, and UTokyo. 
    \item Extensive experimentation of different feature encoders, network components, various augmentation strategies, curriculum learning, and tuning of hyper-parameters.
    
    
   \end{enumerate} 
\brk{What is the motivation of using a GNN based attention network ? the motivation and subsequent usage of these modules in the MuFEm module is unclear even though it seems to be a key contribution}


\section{Related Work} \label{sec:thermal_task}

Usage of thermal imagery for pedestrian detection has received significant interest due to the increasing availability of low-cost thermal sensors. In this section, we first briefly discuss the pedestrian detection literature on visible and thermal sensors. Then we discuss multimodal pedestrian detection based on the fusion of both sensors, which is the focus of this work. We use the more generic term \textit{multimodal} instead of the specific term \textit{multispectral} which refers to thermal and image belong to different spectra.


\subsection{Visible spectrum based pedestrian detection}

Adverse weather conditions such as fog, rain, dust, varied illumination, complex background make the pedestrian detection task extremely difficult in the visible spectrum. To solve such issues to some extent, a joint optimization approach was proposed in \cite{tian2015pedestrian} where semantic attributes were learned for pedestrian detection along with scene attributes. Similar approaches are observed in \cite{ouyang2016learning, brazil2017illuminating} where other vision tasks are added to influence pedestrian detection. In another study \cite{zhang2016faster}, generic Faster-RCNN \cite{ren2015faster} was adapted to solve the problem of pedestrian detection. One of the recent works \cite{liu2019high} proposed an anchor-free method for pedestrian detection.

\subsection{Thermal imagery-based pedestrian detection}

Relatively, there is limited literature on pedestrian detection solely based on thermal imagery. Baek et al. \cite{baek2017efficient} proposed a detection method specific to nighttime only by leveraging thermal-position-intensity-histogram of oriented gradients (T$\pi$HOG) combined with the additive kernel SVM (AKSVM). The method in \cite{ghose2019pedestrian} used thermal images along with saliency maps as part of augmentation. The intention to use this map is to employ an attention mechanism in the network to train Faster R-CNN \cite{ren2015faster}. A ``pseudo-multimodal" object detector was proposed in \cite{devaguptapu2019borrow} where pseudo-RGB equivalent images were generated using Cycle-GAN \cite{zhu2017unpaired} based image-to-image translation framework. Recently, domain adaptation has been applied in \cite{guo2019domain}, \cite{KieuECCV2020taskconditioned} for pedestrian detection to leverage existing camera datasets.

\subsection{Multimodal pedestrian detection}

Fusion of both the visible spectrum and thermal images has been proven to be effective in the recent literature \cite{wagner2016multispectral, liu2016multispectral_, xu2017learning, li2018multispectral}. Yadav et al. \cite{yadavcnn} used a simple fusion strategy and showed good performance for multimodal pedestrian detection.
An extensive study of various fusion architectures was explored in \cite{liu2016multispectral_}. Four different network fusion methods (early, halfway, late, and score fusion) of both visible and thermal data were studied for multimodal pedestrian detection. The halfway fusion model achieved the best detection performance, and we adopt the same fusion strategy. Most of the well-performing multimodal pedestrian detection networks are built upon both variants of R-CNNs. The authors in \cite{li2019illumination} proposed illumination aware Faster R-CNNs to perform multimodal pedestrian detection. Two Single Shot Detectors (SSDs) were employed in \cite{zheng2019gfd} to effectively fuse the features from the visible spectrum and thermal using Gated Fusion Units. Zhang et al. \cite{zhang2019weakly_} highlighted the position shift problem in color-thermal image pairs. This shift in the data for one modality affects another as they are not aligned. The possible reasons for this shift are hardware aging, a shift in field-of-view of two modalities, different resolutions, issues in alignment algorithm, etc. Zhou et al. \cite{zhou2020improving} addressed the modality imbalance problem for multimodal data and presented an in-depth study on the efficient fusion of two modalities. Relatively, there is more literature for LiDAR and image based fusion \cite{zhao2020fusion, el2019rgb}.

\section{Proposed Method} \label{sec:proposed}

\begin{figure*}
    \captionsetup{singlelinecheck=false, font=small, belowskip=-6pt}
    \centering
    \includegraphics[width=\textwidth]{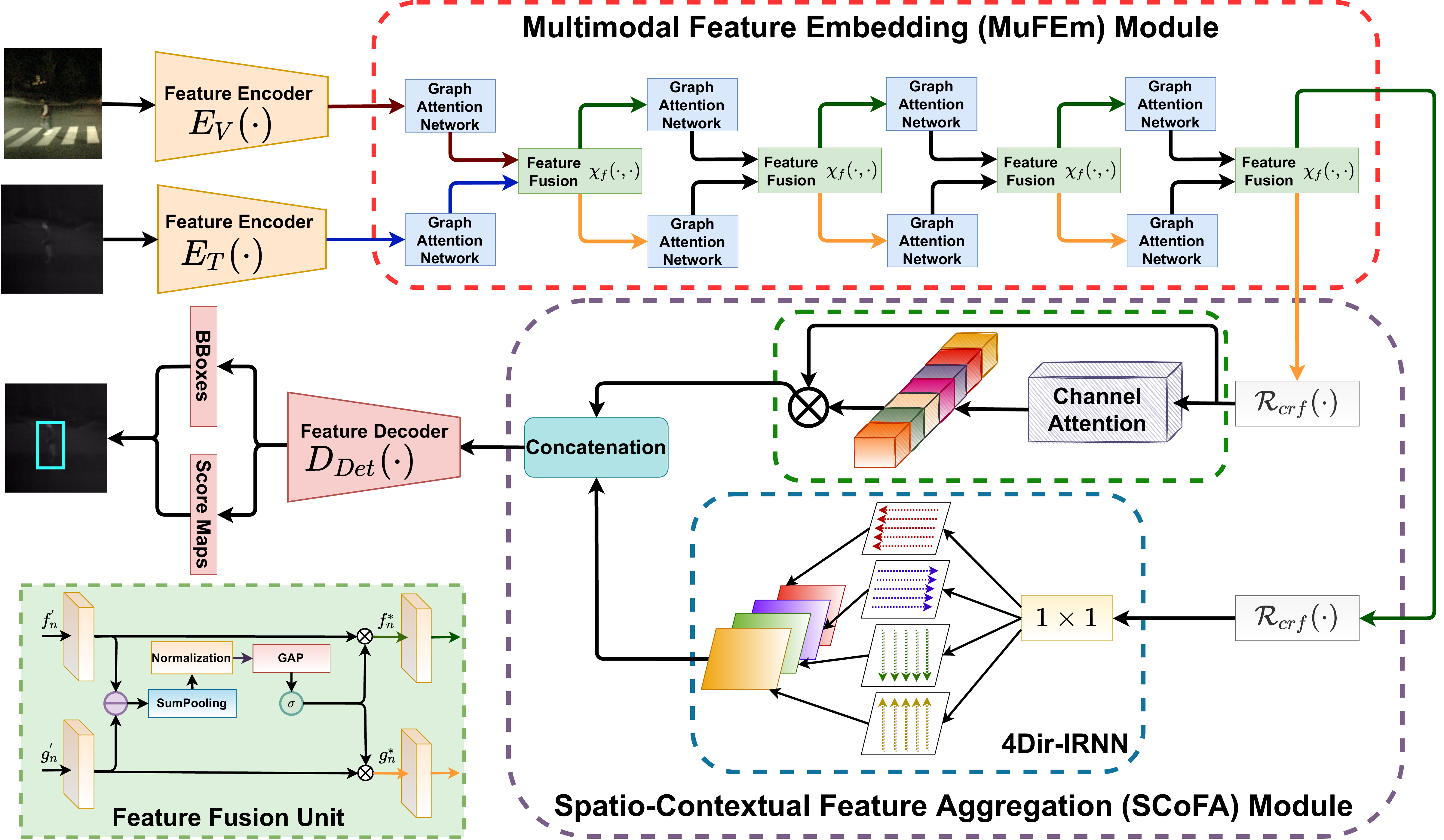} 
    \caption{\textbf{Our proposed architecture for end-to-end multimodal pedestrian detection.}  The input RGB \& thermal images are  passed through two separate deformable ResNext-50 \cite{dai2017deformable} based feature encoders $E_{V}(.)$ \& $E_{T}(.)$ respectively. The outputs from these encoders are passed through a \textit{Multimodal Feature Embedding  (MuFEm) Module} consisting of multiple instances of \textit{Graph Attention Network} \cite{velivckovic2017graph} and  Feature Fusion Unit $\mathcal{X}_{f}(.)$. The two outputs of MuFEM are fed to two distinct CRF based \textit{Feature Aggregation Units} $\mathcal{R}_{crf}(.)$ which belong to a \textit{Spatio-Contextual Feature Aggregation (SCoFA)} Module. The \textit{SCoFA} module additionally contains a channel-wise attention and $4$Dir-IRNN component \cite{bell2016inside} each of which receives input from one of the two $\mathcal{R}_{crf}(.)$. The outputs of these 2 components are concatenated \& passed to the feature decoder $\mathcal{D}_{Det}(.)$ with several deconvolutional layers. The final outputs of $\mathcal{D}_{Det}(.)$ consist of a Score Map \& bounding boxes.}
    \label{fig:proposed_method}
\end{figure*}


The overall architecture of the proposed method is illustrated in Figure \ref{fig:proposed_method}. We use ResNeXt-50 \cite{xie2017aggregated} backbone using Deformable Convolution Network (DCN) \cite{dai2017deformable} to obtain feature extraction independently for both the modalities. The two unimodal encoder feature maps are combined in  \textit{Multimodal Feature Embedding Block (MuFEm)} and further refined in  \textit{Spatio-Contextual Feature Aggregation Block (SCoFA)} to combine the complementary information from the two sensors effectively. Finally, a single stage detection decoder \cite{zhou2017east} uses the multimodal features to output bounding boxes and scores for pedestrians. We discuss these components in detail in the following sub-sections.

\subsection{Unimodal Feature Encoders}

Pedestrians are articulated objects undergoing non-rigid motion. In addition, they suffer from self-occlusion in crowded areas and occlusion due to blockage by other objects. Thus, we are motivated to use deformable convolution, which can handle these geometric deformations better. Deformable convolutions were first introduced in \cite{dai2017deformable} to have an adaptation mechanism in the convolution showing significant improvements over the traditional convolution. FisheyeDistanceNet \cite{kumar2020fisheyedistancenet} successfully made use of it to handle the spatially varying radial distortion. In this work, we use DCN to learn better the appearance of pedestrians with geometric and lighting condition variations. 
Deformable convolution \cite{dai2017deformable} has two parts: a regular convolution layer and 
a deformable grid with a learnable $2$D offset. Grid is adapted based on data, and the convolution layer is applied on the deformed grid. The generalized convolution combining these two operations is shown below.

\begin{equation}
    \mathcal{Y}_{\left(\mathcal{P}_{0}\right)}=\sum_{\mathcal{P}_{i}\in\mathcal{C}}\mathcal{W}(\mathcal{P}_{i})\odot \mathcal{X}(\mathcal{P}_{0}+\mathcal{P}_{i}+\Delta\mathcal{P}_{i})
    \label{eqn:deform_conv}
\end{equation}

where $\mathcal{P}_{0}$ is the center of the grid, and $\mathcal{P}_{i}$ is the point that iterates over the set $\mathcal{C}$ of the sampling points in the grid. Thus the sampling can be irregular and offset locations $\mathcal{P}_{i}$ $+$ $\Delta\mathcal{P}_{i}$, i.e., sampling locations of the kernel are redistributed with the new shapes which may not be rectangular.

\subsection{Multimodal Feature Embedding Block (MuFEm)}

MuFEm uses the unimodal encoder features as the input and outputs multimodal features. {It consists of two parallel blocks of Graph Attention Networks (GAT) \cite{velivckovic2017graph} where the outputs are further fused in the Feature Fusion Unit. This basic unit is repeated several times to obtain progressively aligned feature maps. The details of the GAT and Feature Fusion Unit are discussed in detail below.}

\subsubsection{Graph Attention Network}

Graph Attention Network (GAT) \cite{velivckovic2017graph} was formulated to work on graph data structures using self-attention layers. Their formulation also enabled easy stacking of several GAT layers. In our case, the feature maps are on a dense grid which is a special case of a graph, but it has a very regular structure. There is little work on using graph convolutions for a dense grid. Gao et al. \cite{GaoJ19} extended semantic segmentation using graph convolutions. Cai et al \cite{cai2020remote} successfully applied graph convolutions with cross-attention for remote sensing image classification. We are motivated to use GAT to better handle pixel misalignment across the sensors and non-rigid transformations of pedestrians.

GAT layer uses a graph and its adjacency matrix as the input.
We propose to divide the feature map grid structure $\mathcal{F}_{n}^{'} \in \mathbb{R}^{H\times W\times C}$ into several local feature maps $h= \{ \vec{h_1}, \vec{h_2}, \dots \vec{h_N} \}$, $ \vec{h_i} \in \mathbb{R}^{\mathcal{F}}$ where $N$ is the node of the graph and $\mathcal{F}$ represents the number of features in each node. 
Every node of this graph network is fully connected to each other. Each connection between two nodes captures the relationship between the local features. 
A new set of node features $h= \{ \vec{h^{'}_1}, \vec{h^{'}_2}, \dots \vec{h^{'}_N} \}$, $ \vec{h^{'}_i} \in \mathbb{R}^{\mathcal{F^{'}}}$ is obtained as its output by sharing the information between the adjacency matrix of the previous graph layer nodes and the local features.
A max pooling operation is performed on each extracted local features $\vec{h^{'}_i} \in \mathbb{R}^{\mathcal{F}^{'}} \forall  i $. 
Then a transformation weight matrix which is linear in nature $\mathcal{W} \in \mathbb{R}^{\mathcal{F} \times \mathcal{F^{'}}}$ is applied on each channel to obtain a global feature representation from $\vec{h^{'}_i}$. 
It is followed by a shared self-attention mechanism a: $\mathbb{R}^{\mathcal{F} \times \mathcal{F^{'}}} \rightarrow  \mathbb{R}$  defined in Equation \ref{eqn:self_attention}.
\begin{equation}
    e_{ij}=a(  \mathcal{W}\vec{h_i}, \mathcal{W}\vec{h_j} )
    \label{eqn:self_attention}
\end{equation}

where $i$ and $j$ are two nodes of a graph and $e_{ij}$ is the attention coefficient that captures the information exchange between $i$th and $j$th node. We normalize the function using softmax function in Equation \ref{eqn:softmax_normaliztion} to make the coefficients comparable across different nodes.

\begin{equation}
   \alpha_{ij}= \frac{exp(e_{ij})}{\sum_{k\in\mathcal{N}_{i}}exp(ik)}
    \label{eqn:softmax_normaliztion}
\end{equation}

where $\mathcal{N}_{i}$ is the size of the graph where all the nodes are connected to each other. 

Finally, an attention mechanism based on a feed-forward neural network has been applied, parameterized using a weight vector $ \vec{a}\in \mathbb{R}^{2 \times \mathcal{F}^{'}}$ and employing the LeakyReLU non-linearity ($ \alpha = 0.2$) as shown in the equation below. 
\begin{equation}
  \alpha_{ij}= \frac{exp\left(LeakyReLU(\vec{a^{T}}) [ \mathcal{W}\vec{h_i} \left |  \right |\mathcal{W}\vec{h_j} ]\right)}
  {{\sum_{k\in\mathcal{N}_{i}}}{exp\left(LeakyReLU(\vec{a^{T}})[ \mathcal{W}\vec{h_i} \left |  \right |\mathcal{W}\vec{h_j} ]\right)}}
\end{equation}

\subsubsection{Feature Fusion Unit ($\mathcal{X}_{f}(.)$)} 

The unimodal features modulated by Graph \new{Attention} Network (GAT) are then fed to the Feature Fusion Unit $\mathcal{X}_{f}(.)$. It is designed to enable an effective combination of the unimodal features with the domain gap, mainly based on illumination. It is illustrated in the bottom left corner of Figure \ref{fig:proposed_method}. 

{The Feature Fusion unit receives two sets of input feature $f^{'}_{n}$ and $g^{'}_{n}$ from the two distinct GAT modules. These two sets of feature maps are from two different modalities. We aim to convert the modality-specific features to a modality invariant feature representation for efficient feature embedding reducing the domain gap in feature space. To accomplish this, we subtract the features coming from two different modalities and use them to guide the normalization. Then we feed it into the Global Average Pooling (GAP) layer and apply the \textit{tanh} activation function to obtain modality agnostic feature representation.
Finally, the $f^{'}_{n}$ and $g^{'}_{n}$ are re-calibrated by channel-wise multiplication of domain invariant features obtained. 
We do this iteratively in stages during the course of training that helps to minimize the domain gap and generates domain invariant features for pedestrians.}


\subsection{Spatio-Contextual Feature Aggregation Block (SCoFA)}

\brk{The global architecture of using 2D IRNN is again unclear, what is the motivation in choosing such as architecture ?}
\textit{Spatio-Contextual Feature Aggregation Block (SCoFA)} module consists of two blocks. First, a Conditional Random Fields (CRFs) based aggregation block $\mathcal{R}_{crf}(.)$  followed by channel-wise attention is used to extract the spatial features of the pedestrians. Then, a four-directional IRNN architecture ($4$Dir-IRNN) block \cite{bell2016inside} is used to extract the contextual features of the pedestrians. IRNN is a Recurrent Neural Network (RNN) whose weight matrix is initialized by an Identity matrix.

Liu et al. \cite{liu2019crowd} introduced a deep structured model using CRFs to refine CNN features using a message passing mechanism for the crowd counting problem. Inspired by this work, 
we make use of the  $\mathcal{R}_{crf}(.)$ block to obtain robust and refined features, particularly for crowded pedestrian scenes. There are two distinct $\mathcal{R}_{crf}(.)$ blocks which independently process the two set of feature maps output from \textit{MuFEm} block.
Automotive scenes have a strong context and prior geometry to be exploited. There is a structured roadway for vehicles and a sidewalk for pedestrians. Bell et al \cite{bell2016inside} designed a $4$Dir-IRNN to extract context features for object detection in MS COCO dataset.  We adapt the $4$Dir-IRNN in our proposed architecture to exploit the road scene geometry. Context features can also aid detections in case of occlusions or poor visibility.

In order not to degrade spatial features, we have a parallel $\mathcal{R}_{crf}(.)$ block, which is fed to a channel attention block which is finally concatenated with the context features. The channel-wise attention is used for enhancing inter-channel relationships.
Furthermore, it alleviates the removal of unwanted background information and noise in the feature maps from ${F}_{R_{crf}}$. It is represented by the function below.

\begin{equation}
\mathcal{F}_{att}=\mathcal{F}_{R_{crf}}\otimes\sigma   (\mathcal{C}onv_{1\times1}(\mathcal{F}_{R_{crf}}))
\end{equation}

Each instance of RNN consumes a sequence of $512$ channels of the input feature map after being processed by a  {$\mathcal{C}onv_{1\times1}$} block. $4$Dir-IRNN traverses in four directions (right to the left, left to right, top to bottom, and bottom to top) illustrated in Figure \ref{fig:proposed_method} to obtain context features.  Furthermore, the contextual features obtained from the multi-directional RNNs are directly concatenated to produce {$\mathcal{F}_{IRNN}$} which is further concatenated to $\mathcal{F}_{att}$ to obtain the final feature vector.

\subsection{Feature Decoder Block \& Detection Branch}

We design a single-stage detection decoder $\mathcal{D}_{Det}(.)$ which uses the feature maps from the \textit{ScoFA} as input and outputs bounding boxes with score maps.

We adapt the EAST decoder \cite{zhou2017east} designed for efficient text detection. We progressively upsampled the previous layer by $2$ and reduced the number of channels from $256$ to $32$ before feeding into the detection branch. Then it is passed through four {$\mathcal{C}onv_{1\times1}$} layers that produce $4$ channels of feature maps. The final layer regresses the score map and rotated boxes.
The loss function of detector is provided in Equation \ref{eqn:loss}. 

\begin{equation}
\mathcal{L} = \mathcal{L}_{s}+\mathcal{\lambda}_{g}\mathcal{L}_{g}, \label{eqn:loss}
\end{equation}

where $\mathcal{L}_{s}$ and $\mathcal{L}_{g}$ correspond to the loss of the Score Map and geometric loss of rotated box, respectively, $ \mathcal{\lambda}_{g}$ represents the weighting ratio between the two losses.

\begin{table*}[t]
    \centering
    {
    
    \begin{tabular}{lcccccccccccccccccc}
    \toprule
        \textbf{Encoders} & \multicolumn{5}{c}{\textbf{Number of Fusion Units}} & \multicolumn{3}{c}{\textbf{Feature Aggregation}} &  \multicolumn{3}{c}{\textbf{Miss Rate}} \\
        \cmidrule(r){2-6}
        \cmidrule(r){7-9}
        \cmidrule(r){10-12}
        & \textbf{1} & \textbf{2}  & \textbf{3} & \textbf{4} & \textbf{5} &  \textbf{Spatial} & \textbf{Contextual} &  \textbf{Both} & \textbf{All} & \textbf{Day} & \textbf{Night}\\
         \cmidrule(r){1-12} 
         
    \multirow{15}{*}{ResNet-50 \cite{he2016deep}} &  \color{green}\Checkmark & \color{red}\xmark  & \color{red}\xmark & \color{red}\xmark & \color{red}\xmark & \color{green}\Checkmark & \color{red}\xmark & \color{red}\xmark & 26.23 & 27.35 & 25.67 \\

  & \color{green}\Checkmark & \color{red}\xmark & \color{red}\xmark & \color{red}\xmark & \color{red}\xmark & \color{red}\xmark  & \color{green}\Checkmark & \color{red}\xmark & 25.01 & 25.17 &  24.45\\
   
    & \color{green}\Checkmark & \color{red}\xmark & \color{red}\xmark & \color{red}\xmark & \color{red}\xmark & \color{red}\xmark  & \color{red}\xmark  & \color{green}\Checkmark & 23.98 &  24.66 & 22.01 \\
   
  \cmidrule(r){2-12}

  & \color{red}\xmark  & \color{green}\Checkmark  & \color{red}\xmark & \color{red}\xmark & \color{red}\xmark & \color{green}\Checkmark & \color{red}\xmark & \color{red}\xmark & 23.53 & 23.62   & 21.97 \\

  & \color{red}\xmark & \color{green}\Checkmark  & \color{red}\xmark & \color{red}\xmark & \color{red}\xmark & \color{red}\xmark  & \color{green}\Checkmark & \color{red}\xmark & 22.31  & 23.99 & 20.78 \\
   
    & \color{red}\xmark & \color{green}\Checkmark & \color{red}\xmark & \color{red}\xmark & \color{red}\xmark & \color{red}\xmark  & \color{red}\xmark  & \color{green}\Checkmark  & 20.87 & 20.96 & 19.31 \\

  \cmidrule(r){2-12}

 & \color{red}\xmark  & \color{red}\xmark  & \color{green}\Checkmark & \color{red}\xmark & \color{red}\xmark & \color{green}\Checkmark & \color{red}\xmark & \color{red}\xmark & 21.83 & 21.97 & 21.27 \\

  & \color{red}\xmark & \color{red}\xmark  & \color{green}\Checkmark & \color{red}\xmark & \color{red}\xmark & \color{red}\xmark  & \color{green}\Checkmark & \color{red}\xmark  & 20.61 & 20.74 & 20.05 \\
   
    & \color{red}\xmark & \color{red}\xmark  & \color{green}\Checkmark  & \color{red}\xmark & \color{red}\xmark & \color{red}\xmark  & \color{red}\xmark  & \color{green}\Checkmark  & 18.17 & 19.26 & 19.11 \\

\cmidrule(r){2-12}

 & \color{red}\xmark  & \color{red}\xmark  & \color{red}\xmark & \color{green}\Checkmark  & \color{red}\xmark & \color{green}\Checkmark & \color{red}\xmark & \color{red}\xmark & 19.13 & 19.22 & 18.57\\

  & \color{red}\xmark & \color{red}\xmark  & \color{red}\xmark  & \color{green}\Checkmark & \color{red}\xmark & \color{red}\xmark  & \color{green}\Checkmark & \color{red}\xmark  & 18.91 & 19.17 & 18.35\\
   
    & \color{red}\xmark & \color{red}\xmark  & \color{red}\xmark  & \color{green}\Checkmark  & \color{red}\xmark & \color{red}\xmark  & \color{red}\xmark  & \color{green}\Checkmark & 15.47 &  16.56 & 15.91 \\

\cmidrule(r){2-12}

& \color{red}\xmark  & \color{red}\xmark  & \color{red}\xmark &\color{red}\xmark  &  \color{green}\Checkmark& \color{green}\Checkmark & \color{red}\xmark & \color{red}\xmark & 19.09 & 19.2 & 18.35\\

  & \color{red}\xmark & \color{red}\xmark  & \color{red}\xmark  & \color{red}\xmark &  \color{green}\Checkmark& \color{red}\xmark  & \color{green}\Checkmark & \color{red}\xmark & 18.78 & 19.12 & 18.16\\
   
    & \color{red}\xmark & \color{red}\xmark  & \color{red}\xmark  & \color{red}\xmark   &  \color{green}\Checkmark& \color{red}\xmark  & \color{red}\xmark  & \color{green}\Checkmark & 15.26 & 16.44 & 16.21 \\


\midrule
    
    \multirow{15}{*}{\centering{ResNeXt-50 \cite{xie2017aggregated}}}  &  \color{green}\Checkmark & \color{red}\xmark & \color{red}\xmark & \color{red}\xmark & \color{red}\xmark & \color{green}\Checkmark & \color{red}\xmark & \color{red}\xmark & 20.53 & 20.62 & 19.97 \\

  & \color{green}\Checkmark & \color{red}\xmark & \color{red}\xmark & \color{red}\xmark & \color{red}\xmark & \color{red}\xmark  & \color{green}\Checkmark & \color{red}\xmark &  18.76 & 19.08 & 19.64 \\
   
    & \color{green}\Checkmark & \color{red}\xmark & \color{red}\xmark & \color{red}\xmark & \color{red}\xmark & \color{red}\xmark  & \color{red}\xmark  & \color{green}\Checkmark & 17.87  & 18.96 & 17.31\\
   
  \cmidrule(r){2-12}
   
  & \color{red}\xmark  & \color{green}\Checkmark  & \color{red}\xmark & \color{red}\xmark & \color{red}\xmark & \color{green}\Checkmark & \color{red}\xmark & \color{red}\xmark & 16.02 & 17.11  & 16.46 \\

  & \color{red}\xmark & \color{green}\Checkmark  & \color{red}\xmark & \color{red}\xmark & \color{red}\xmark & \color{red}\xmark  & \color{green}\Checkmark & \color{red}\xmark & 15.88   &  16.97 &  16.22\\
   
    & \color{red}\xmark & \color{green}\Checkmark & \color{red}\xmark & \color{red}\xmark & \color{red}\xmark & \color{red}\xmark  & \color{red}\xmark  & \color{green}\Checkmark &  13.67 & 15.76 & 14.11 \\

  \cmidrule(r){2-12}

 & \color{red}\xmark  & \color{red}\xmark  & \color{green}\Checkmark & \color{red}\xmark & \color{red}\xmark & \color{green}\Checkmark & \color{red}\xmark & \color{red}\xmark & 13.62 & 14.73 & 14.66\\

  & \color{red}\xmark & \color{red}\xmark  & \color{green}\Checkmark & \color{red}\xmark & \color{red}\xmark & \color{red}\xmark  & \color{green}\Checkmark & \color{red}\xmark & 11.48 &  13.03 & 12.8\\
   
    & \color{red}\xmark & \color{red}\xmark  & \color{green}\Checkmark  & \color{red}\xmark & \color{red}\xmark & \color{red}\xmark  & \color{red}\xmark  & \color{green}\Checkmark & 10.27  & 11.96 & 11.41 \\

\cmidrule(r){2-12}

 & \color{red}\xmark  & \color{red}\xmark  & \color{red}\xmark & \color{green}\Checkmark  & \color{red}\xmark & \color{green}\Checkmark & \color{red}\xmark & \color{red}\xmark & 11.52 & 12.87 & 11.92\\

  & \color{red}\xmark & \color{red}\xmark  & \color{red}\xmark  & \color{green}\Checkmark & \color{red}\xmark & \color{red}\xmark  & \color{green}\Checkmark & \color{red}\xmark & 10.38 & 10.49 & 9.93\\
   
    & \color{red}\xmark & \color{red}\xmark  & \color{red}\xmark  & \color{green}\Checkmark  & \color{red}\xmark & \color{red}\xmark  & \color{red}\xmark  & \color{green}\Checkmark &  \textbf{9.23} & \textbf{9.33} &  \textbf{8.97} \\

\cmidrule(r){2-12}

& \color{red}\xmark  & \color{red}\xmark  & \color{red}\xmark &\color{red}\xmark  &  \color{green}\Checkmark& \color{green}\Checkmark & \color{red}\xmark & \color{red}\xmark &11.56 & 12.67 & 12.03 \\

  & \color{red}\xmark & \color{red}\xmark  & \color{red}\xmark  & \color{red}\xmark &  \color{green}\Checkmark& \color{red}\xmark  & \color{green}\Checkmark & \color{red}\xmark  & 10.45 & 10.57 & 10.0\\
   
    & \color{red}\xmark & \color{red}\xmark  & \color{red}\xmark  & \color{red}\xmark   &  \color{green}\Checkmark& \color{red}\xmark  & \color{red}\xmark  & \color{green}\Checkmark & 9.39 & 9.74 & 9.06\\
    \bottomrule
    \end{tabular}}
    \caption{\textbf{Ablation study of the proposed architecture on KAIST dataset \cite{hwang2015multispectral}.}  \textit{MuFEm} module consists of several cascaded blocks of Graph Attention Network \cite{velivckovic2017graph} \&  a Feature Fusion Unit.  
    Ablation study is conducted on spatial (channel wise attention) \& 
    contextual ($4$Dir-IRNN \cite{bell2016inside}) feature branch of \textit{SCoFA} modules varying the number of  Fusion Units from $1$ to $5$.} 
    \label{tab:arch_ablation}
    \vspace{-7mm}
\end{table*}

Bounding box is represented by the top left and right bottom ends represented in the output coordinate grid as $p_{t}=(x_{t}, y_{t})$ and $p_{b}=(x_{b}, y_{b})$. For every pixel, the output feature map consists of a bounding box and a detection score. We use the class balanced cross-entropy loss \cite{xie2015holistically} in score map prediction as shown in Equation \ref{Eqn:Score_Map}.

\begin{equation}
\mathcal{L}_{s} = -\beta S^{GT} log S^{P} - (1- \beta)(1-S^{GT})log(1-S^{p})
\label{Eqn:Score_Map}
\end{equation}

where $S^{P}$ is the prediction of the score map and ${S^{GT}}$ is ground truth. 
It is to be noted that $\beta$ plays the role of class balancing weight factor between positive and negative input samples. 

The sizes of pedestrians vary widely and it is a main challenge in the pedestrian detection problem. Regression loss using $L1$ or $L2$ may give more importance to larger sized bounding boxes and a scale-invariant loss is more suitable. Thus we chose to use the IoU loss \cite{zhou2017east} for the Bounding Box (BBox) regression as shown below. 
\begin{equation}  \label{Eqn:BBox}
\begin{aligned}
     \mathcal{L}_{g} &= -log IoU(BBox^{P}, BBox^{GT}) \\ 
     &= - log \frac{\left |  BBox^{P} \cap BBox^{GT}\right |}{\left |  BBox^{P} \cup  BBox^{GT}\right |}
\end{aligned}    
\end{equation}

where $BBox^{P}$ is the predicted bounding boxes of the pedestrians while $BBox^{GT}$ represents the ground truth of the corresponding pedestrians.

\section{Experimentation Details} 
\label{sec:performance}

\definecolor{LightCyan}{rgb}{0.88,1,1}
\definecolor{LightGreen}{rgb}{0.35,0.8,0}
\definecolor{LightGreen2}{rgb}{0.5,0.9,0}
\definecolor{Gray}{gray}{0.85}

\subsection{Dataset and evaluation metrics}

In this work, we make use of the following three main publicly available multimodal datasets discussed below. 

\textbf{KAIST:} KAIST \cite{hwang2015multispectral} is the commonly used dataset for
Multimodal Pedestrian Detection.  It consists of $95,000$ \textit{color-thermal} pairs of frames with $103,128$ annotated bounding box and $1,182$ distinct pedestrians. The frames were captured in both day and night time scenarios and it contains more granular annotations such as sunrise, morning, afternoon, sunset, night, and dawn. The dataset contains pedestrians with many partially occluded and heavily occluded instances. 
Alignment issues in the original dataset were fixed by Zhang et al. \cite{zhang2019weakly_} and test annotations were refined by Liu et al. \cite{liu2016multispectral_}. We use the latest version of the dataset with these fixes. 

\textbf{CVC-14:} The CVC-14 dataset \cite{gonzalez2016pedestrian} contains video sequences of \textit{grayscale-thermal} pair captured at $10$ FPS during day and night time. The sample distribution for training and testing in the dataset follows $7,085$ and $1,433$ frames, respectively. Annotations for each modality are provided individually as the camera extrinsic was not calibrated properly. 

\textbf{UTokyo:} The UTokyo multimodal dataset \cite{takumi2017multispectral} contains a total of $7,512$ images captured during day ($3,740$) and night ($3,772$) time at $1$ FPS in a university campus using RGB, far-infrared (FIR), mid-infrared (MIR), and near-infrared (NIR) cameras. In this work, we only use the samples that are aligned to enable comparison with other methods.

For detection accuracy of the pedestrians, we use the standard log average Miss Rate (MR) to calculate the error. We also use mAP (mean average precision) to compare with few other works. Note that higher mAP is better, whereas lower MR is better.
All methods reported in this work are evaluated based on the standard protocol called \textit{reasonable} setup \cite{dollar2011pedestrian} for KAIST dataset \cite{hwang2015multispectral} where pedestrian boxes are over 50 pixels tall under no or partial occlusion. \new{Our models were designed and trained from scratch without any use of pretrained models. The model is independently trained and tested on each dataset. It uses only one frame as input at both training and inference time and does not exploit temporal information.}

\subsection{Ablation study}

We chose KAIST \cite{hwang2015multispectral} for our ablation studies as it is the largest and the most commonly used dataset for multimodal pedestrian detection. \new{We use the test set to report the results.} Thus, we perform a broad range of systematic experiments as part of the ablation study on this dataset to find the optimal configuration of our model. The network is trained using Momentum Optimizer with an initial learning rate of $0.01$ and momentum set to $0.9$. A factor of $10$ then reduces the learning rate after the completion of every $5$k iterations. The proposed model was trained on a system with two Nvidia Tesla P6 GPUs, and the batch size was set to $1$. \new {Our end-to-end detection framework runs at 10 fps on the Nvidia Tesla P6 GPU.}

\textbf{Network components:} The basic fusion unit in \textit{MuFEm} module consists of a pair of Graph Attention Networks (GATs) operating parallelly, and a feature fusion unit combines their outputs. We repeat these blocks to combine the unimodal feature maps progressively. We vary the number of fusion units from $1$ to $5$. For each of these configurations, we do ablation of the spatial and contextual components of the \textit{SCoFA} module. Experiments are performed with both these enabled together and with only one of them enabled. To keep the size of the network reasonable, we make use of ResNet-$50$ \cite{he2016deep} as the baseline instead of ResNet-$101$. We also used ResNeXt-$50$ \cite{xie2017aggregated} encoder which is an improvement over ResNet.
Both these encoders used deformable convolution \cite{dai2017deformable} as our initial set of experiments demonstrate a significant improvement over regular convolutions. Thus we made an early design choice to use DCN and we did not perform ablation study on this with other network components to limit the number of combinations.

Table \ref{tab:arch_ablation} summarizes the ablation study of network architecture components. Accuracy increases progressively with number of fusion units and saturates at $4$. Spatial and contextual blocks in \textit{SCoFA} provide a significant improvement on top of optimal \textit{MuFEm} setting. The best setting is for usage of both spatial and contextual blocks in \textit{SCoFA} and $4$ fusion units in \textit{MuFEm} with ResNeXt-$50$ encoder. All the further experiments make use of this  configuration.


\textbf{Curriculum  Learning:} 
Curriculum Learning \cite{bengio2009curriculum} is a training method that aims to gradually increase the complexity of the data fed to the input neural network.  The main idea is to train using easier objects first and gradually increase the object's difficulty level. It provided stable convergence to the global optimum.
Mask and Predict \cite{kishore2019cluenet} is a specific strategy in curriculum learning where the pedestrian box is progressively masked, and the network is expected to predict the boxes with the visible and masked regions. It was used to handle partially visible pedestrians due to occlusion. In this paper, we use the mask and predict-based curriculum learning defined in our previous work \cite{das2020end}. We mask the pedestrian regions with masking $\%$  increasing from $0\%$ to $70\%$ gradually over the training phases. As per Table \ref{tab:ablation_data_aug}, we can observe that curriculum learning provides a significant improvement even on top of two data augmentation techniques. Our qualitative results show a good performance in occluded scenes and an improvement in generalization.


\begin{table}[t]

\centering

\scalebox{1}{
\begin{tabular}{c|c|c|c|c|c} 
\bottomrule
\multirow{2}{*}{\cellcolor{blue!25}} & \multicolumn{2}{c}{ $\textbf{\cellcolor{yellow!25}Data Aug.}$} & \multirow{2}{*}{\cellcolor{red!25}} & \multirow{2}{*}{\cellcolor{red!25}} & \multirow{2}{*}{\cellcolor{red!25}}\\

\cline{2-3}
\multirow{-2}{*}{\cellcolor{blue!25}$\textbf{\makecell{Curriculum\\Learning}}$}& \cellcolor{yellow!25}$\textbf{S}$ & \cellcolor{yellow!25}$\textbf{M}$ & \multirow{-2}{*}{\cellcolor{red!25}$\textbf{MR All}$$\downarrow$} & \multirow{-2}{*}{\cellcolor{red!25}$\textbf{MR Day}$$\downarrow$} & \multirow{-2}{*}{\cellcolor{red!25}$\textbf{MR Night}$$\downarrow$}\\

\midrule
\rowcolor{LightCyan}
\multicolumn{6}{c}{\textbf{Multimodal}}
\\
\midrule

\color{green}\Checkmark & \color{red}\xmark & \color{red}\xmark & 9.1 & 9.19 & 8.93\\ 

\color{green}\Checkmark & \color{green}\Checkmark & \color{red}\xmark & 8.6 & 9.0 & 8.51\\  

\color{green}\Checkmark & \color{red}\xmark & \color{green}\Checkmark & 8.54 & 8.92 & 8.47\\ 

\color{red}\xmark & \color{green}\Checkmark & \color{red}\xmark & 9.17 & 9.23 & 8.81\\ 

\color{red}\xmark & \color{red}\xmark & \color{green}\Checkmark & 9.03 & 9.18 & 8.87\\

\color{red}\xmark & \color{green}\Checkmark & \color{green}\Checkmark & 8.83 & 9.01 & 8.6\\

\color{green}\Checkmark & \color{green}\Checkmark & \color{green}\Checkmark & \textbf{8.31} & \textbf{8.64} & \textbf{8.0}\\

\bottomrule


\end{tabular}

}
\caption{\textbf{Ablation study of different training strategies  on KAIST dataset \cite{hwang2015multispectral}. } S and M stand for simple and mixup augmentation respectively.}
\label{tab:ablation_data_aug}
\end{table}

\begin{table}[t]
\centering

\scalebox{1}{
\begin{tabular}{c|c|c|c}
\bottomrule
\cellcolor{blue!25} $\textbf{\textbf{$\lambda_g$}}$  & \cellcolor{red!25}$\textbf{MR All}$$\downarrow$  & \cellcolor{red!25}$\textbf{MR Day}$$\downarrow$ & \cellcolor{red!25}$\textbf{MR Night}$$\downarrow$  \\

\midrule

\rowcolor{LightCyan}
\multicolumn{4}{c}{\textbf{Multimodal}}
\\
\midrule

$0.1$ & 8.42 & 8.86 & 8.2\\ 

$0.3$ & 8.31 & 8.64 & 8.0\\ 

$0.7$ & 8.19 & 8.30 & 7.76 \\ 

\textbf{1.0} & \textbf{8.07}  & \textbf{8.16} & \textbf{7.51} \\ 

\bottomrule
\end{tabular}
}
\caption{\textbf{Comparison of different loss balancing factors ($\lambda_g$) in Eqn. \ref{eqn:loss} on KAIST dataset\cite{hwang2015multispectral}. } MR is Miss Rate.
}

\vspace{2mm}
    
\label{tab:hyperparameters}

\end{table}

    
     

\begin{table*}[!h]
\centering

\scalebox{1}{
\begin{tabular}{l|c|c|c|c|c|c|c|c|c} 
\bottomrule
  & \multicolumn{3}{c|} {\textbf{Multimodal (MR $\downarrow$)}} & \multicolumn{3}{c|}{\textbf{Visible (MR $\downarrow$)}} & \multicolumn{3}{c|}{\textbf{Thermal (MR $\downarrow$)}}\\
\cline{1-10}
\cellcolor{blue!25} $\textbf{Methods}$  & \cellcolor{red!25}$\textbf{All}$  & \cellcolor{red!25}$\textbf{Day}$ & \cellcolor{red!25}$\textbf{Night}$ & \cellcolor{red!25}$\textbf{All}$  & \cellcolor{red!25}$\textbf{Day}$ & \cellcolor{red!25}$\textbf{Night}$ & \cellcolor{red!25}$\textbf{All}$  & \cellcolor{red!25}$\textbf{Day}$ & \cellcolor{red!25}$\textbf{Night}$  \\
\toprule

ACF \cite{hwang2015multispectral} & 31.34 & 29.59 & 34.98 & 32.01 & 29.85 & 36.77 & 31.90 & 30.40 & 34.81\\

Halfway Fusion  \cite{liu2016multispectral_} & 25.75  & 24.88 & 26.59 & 25.10  & 24.29 & 26.12 & 25.51  & 25.20 & 24.90  \\ 

Fusion RPN  \cite{konig2017fully} & 20.67  & 19.55 & 22.12 & 20.52  & 19.69 & 21.83 & 21.43  & 21.08 & 20.88 \\ 

Fusion RPN+BF  \cite{konig2017fully} & 15.91  & 16.49 & 15.15 & 15.98  & 16.60 & 15.28 & 16.52  & 17.56 & 14.48 \\ 

IAF R-CNN  \cite{li2019illumination} & 15.73  & 14.55 & 18.26 & 15.65  & 14.95 & 18.11 & 16.00  & 15.22 & 17.56 \\ 

IATDNN+IASS  \cite{guan2019fusion} & 14.95  & 14.67 & 15.72 & 15.14  & 14.82 & 15.87 & 15.08  & 15.02 & 15.20\\ 

RFA \cite{zhang2019cross} & 14.61 & 16.78 & 10.21 & --  & -- & -- & --  & -- & --\\

CIAN \cite{zhang2019cross_j} & --  & -- & -- & 14.64 & 15.13 & 12.43 & 14.68 & 16.21 & 9.88 \\

MSDS-RCNN  \cite{li2018multispectral} & 11.63  & 10.60 & 13.73 & 11.28  & 9.91 & 14.21  & 12.51  & 12.02 & 13.01  \\ 

CS-RCNN \cite{zhang2020attention} & 11.43 & 11.86 & 8.82 & --  & -- & -- & --  & -- & -- \\

AR-CNN \cite{zhang2019weakly_} & 9.34 & 9.94 & 8.38 & 8.86 & 8.45 & 9.16 & 8.26 & 9.02 & 7.04 \\

MBNet \cite{zhou2020improving} & 8.13 & 8.28 & 7.86 & --  & -- & -- & --  & -- & --\\

\rowcolor{LightGreen}
\textbf{Ours}  & \textbf{8.07} & \textbf{8.16} & \textbf{7.51} & \textbf{11.79} & \textbf{9.89} & \textbf{14.04} & \textbf{8.18} & \textbf{9.29} & \textbf{7.03}\\
\toprule

\end{tabular}
}
\caption{\textbf{Comparison of multimodal pedestrian detection on KAIST  dataset \cite{hwang2015multispectral} using Miss Rate (MR) metric.}}
\label{tab:detection_multimodality_kaist}

\vspace{-3mm}

\end{table*}

\begin{table}[t]
\centering

\scalebox{1}{
\begin{tabular}{l|c} 
\bottomrule
\cellcolor{blue!25}\textbf{Methods}  & \cellcolor{red!25}$\textbf{mAP}$ $\uparrow$    \\
\toprule

PiCA-Net  \cite{ghose2019pedestrian} & 65.80*  \\

R$^3$Net  \cite{ghose2019pedestrian} & 70.85*   \\

tY  \cite{krivsto2020thermal} & 63.00   \\

SSTN101  \cite{munir2021sstn} & 73.22   \\

\rowcolor{LightGreen}
\textbf{Ours}  & \textbf{78.03}  \\
\toprule

\end{tabular}
}
\caption{\textbf{
Quantitative comparison of multimodal pedestrian detection on KAIST  dataset \cite{hwang2015multispectral} using mAP metric.} (*) denotes the average of day and night mAP scores.
}

\vspace{4mm}

\label{tab:detection_multimodality_kaist_map}

\end{table}

\textbf{Data augmentation:} Data augmentation is an essential ingredient of deep learning-based object detection algorithms to regularize the model. In this work, we apply the standard augmentation techniques such as random scaling of images between $0.8$ to $1.2$, image flipping with a probability of $0.3$, enhancing contrast, adding Gaussian noise, and blur effect. We refer to these as simple augmentations ($S$). In addition to this, we use $mixup$ \cite{zhang2017mixup}, a data-agnostic augmentation technique that trains a network on pairs of examples and their labels based on convex combinations.  $mixup$ $(M)$ can be represented as below.

\begin{equation}\label{eqn:mixup}
\begin{aligned}
  \tilde{m} &= \mathcal{\omega}m_{i}+(1- \mathcal{\omega})m_{j}\\
  \tilde{n} &= \mathcal{\omega} n_{i}+(1- \mathcal{\omega})n_{j}
\end{aligned}
\end{equation}

where $m_{i}, m_{j}$ are raw input vectors and $n_{i}, n_{j}$ are one-hot label encoding. $(m_{i},n_{i})$ and $(m_{j},n_{j})$ are two examples generated randomly from the training dataset and $\mathcal{\omega} \in [0,1]$. This technique is simple to implement and has low computational overhead. It provides regularization to noisy labels by generating mixed up patches. From Table \ref{tab:ablation_data_aug}, it can be observed $mixup$ provides an additional improvement on top of $S$ with or without curriuculum learning.

\textbf{Loss balancing factor:} The hyperparameter $\lambda_g$ denotes the loss balancing factor between geometric loss ($\mathcal{L}_{g}$) and score map loss ($\mathcal{L}_{s}$) as shown in Equation \ref{eqn:loss}. The range of this value is [$0,1$] and we heuristically test with different predefined values ($0.1, 0.3, 0.7$ and $1.0$) of $\lambda_g$. MRs are reported in  Table \ref{tab:hyperparameters} showing that the optimal value of $\lambda_g$ is 1.

\subsection{Quantitative study}
To facilitate a quantitative comparison with the state-of-the-art techniques, we evaluate our method on KAIST \cite{hwang2015multispectral}, CVC-14 \cite{gonzalez2016pedestrian} and UTokyo \cite{takumi2017multispectral} multimodal datasets.


\begin{figure*}[t]
    \centering
    \includegraphics[width=0.245\textwidth]{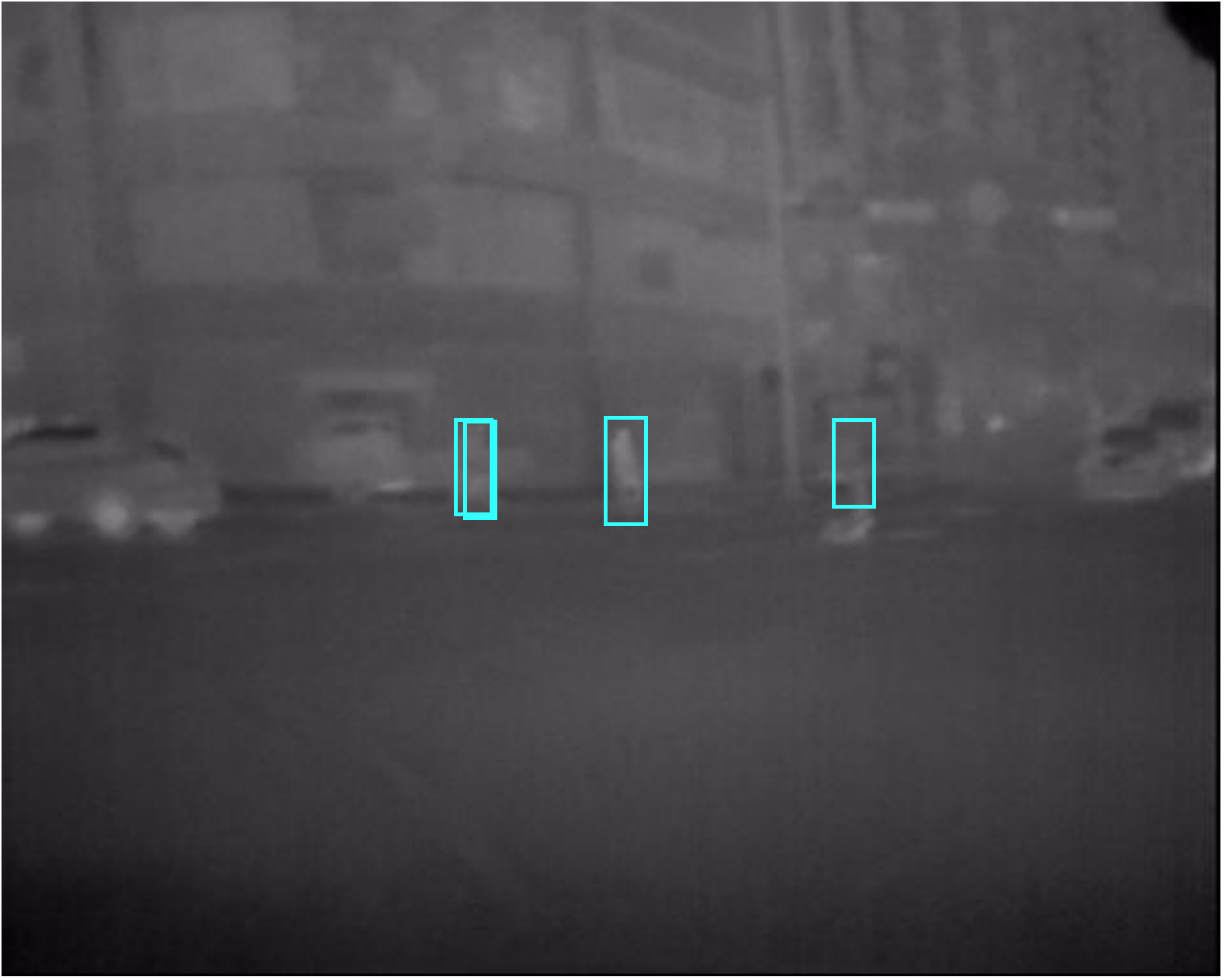}
    \includegraphics[width=0.245\textwidth]{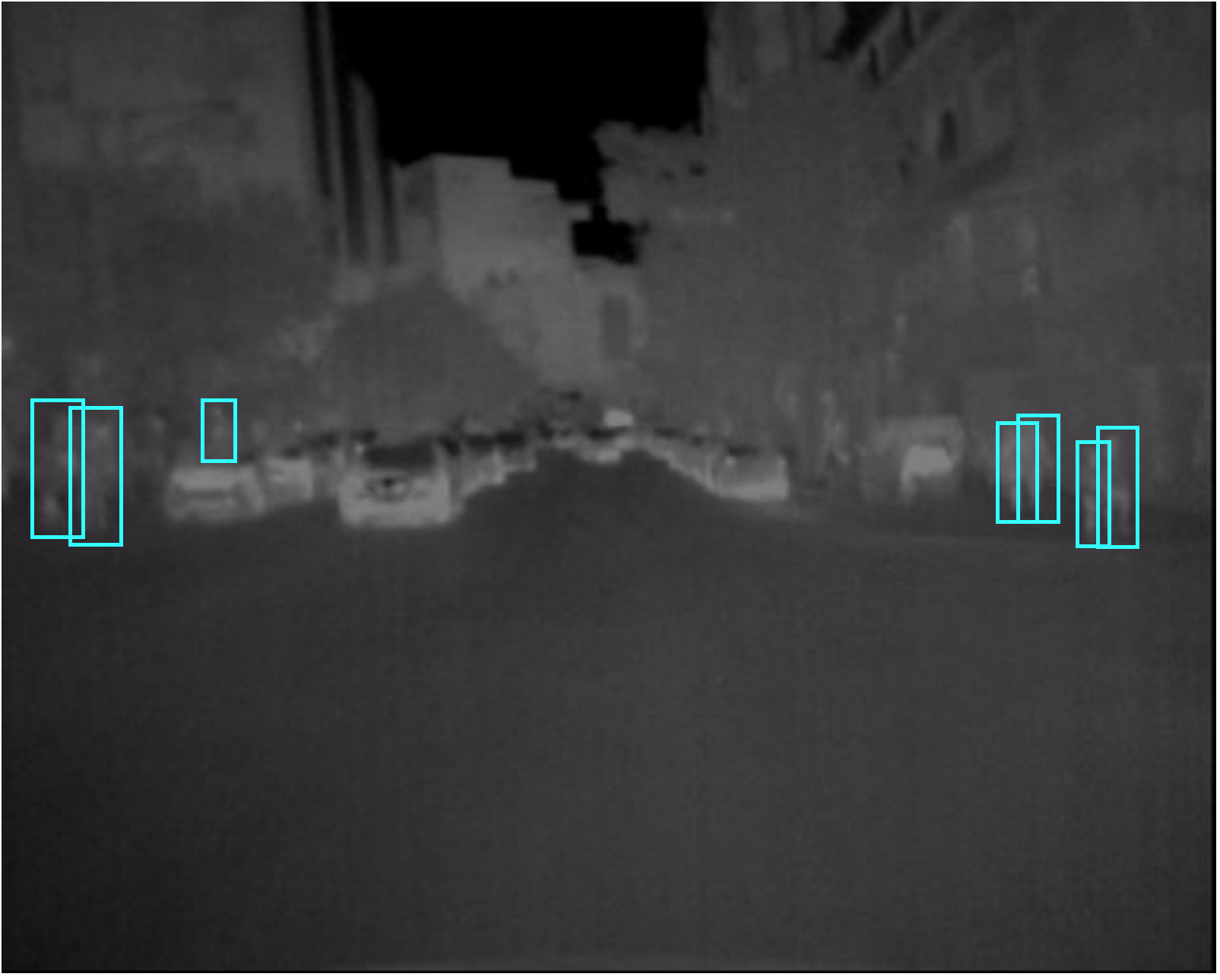}
    \includegraphics[width=0.245\textwidth]{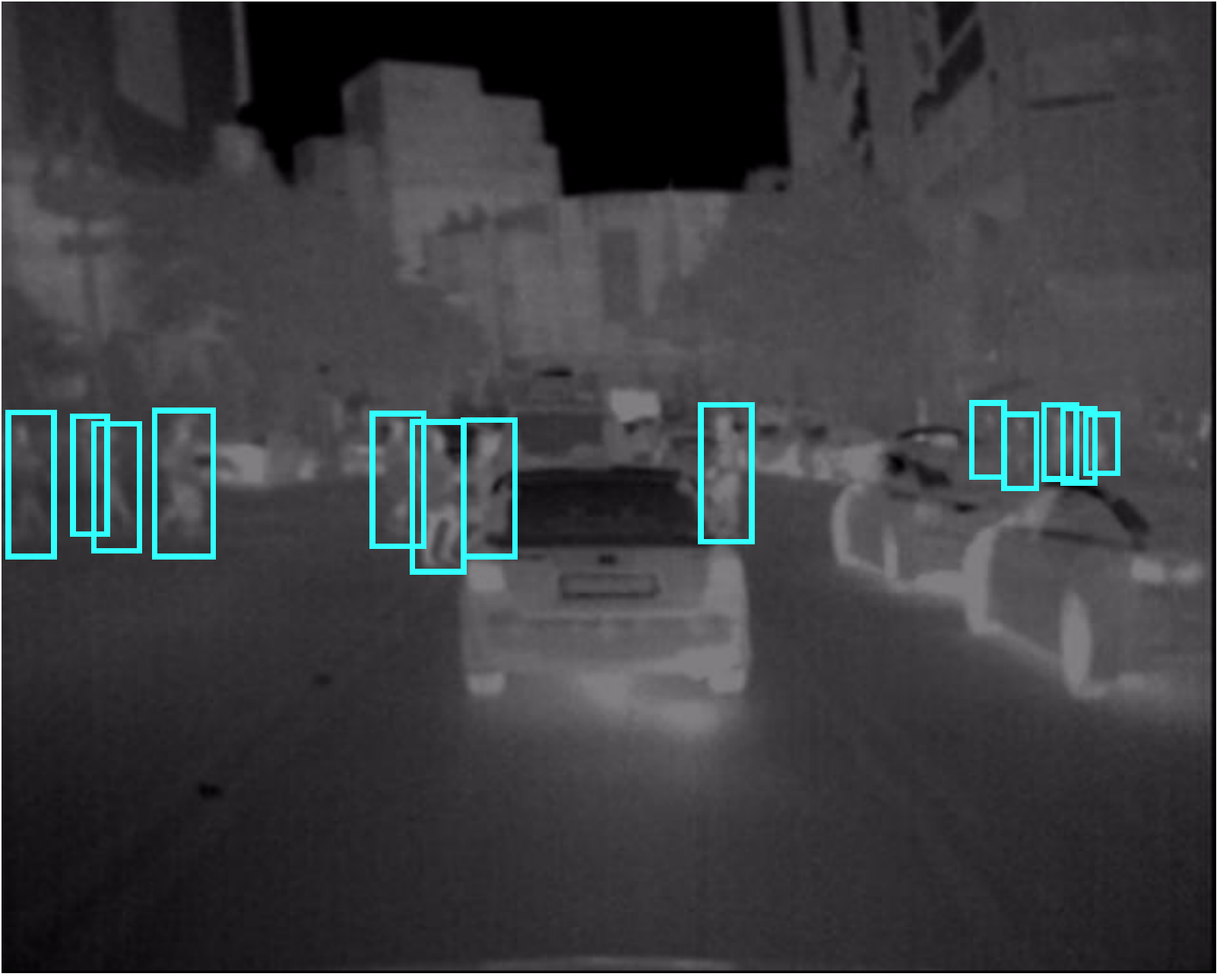}
    \includegraphics[width=0.245\textwidth]{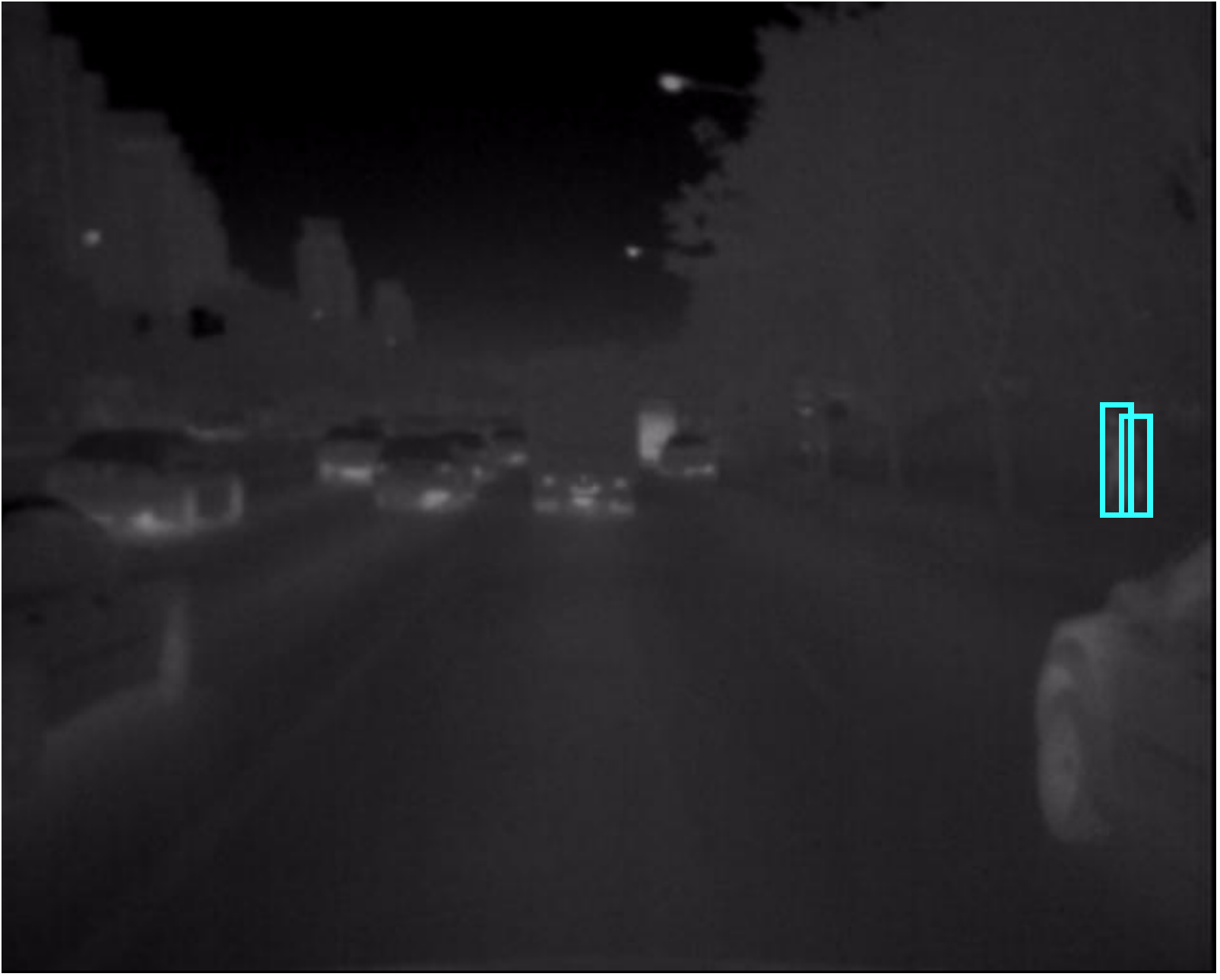} 
    
     \vspace{0.25 mm}
     
    \includegraphics[width=0.245\textwidth]{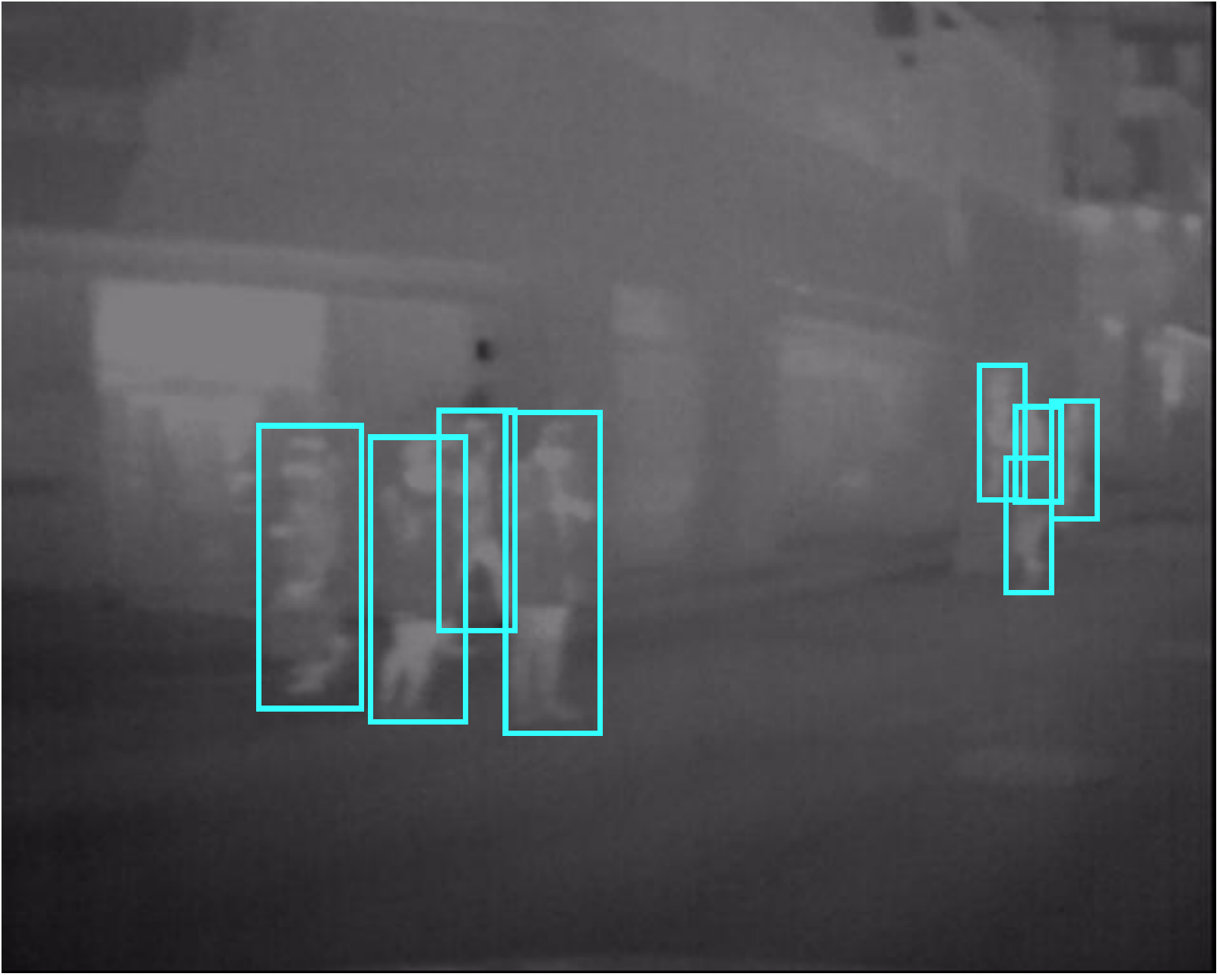}
    \includegraphics[width=0.245\textwidth]{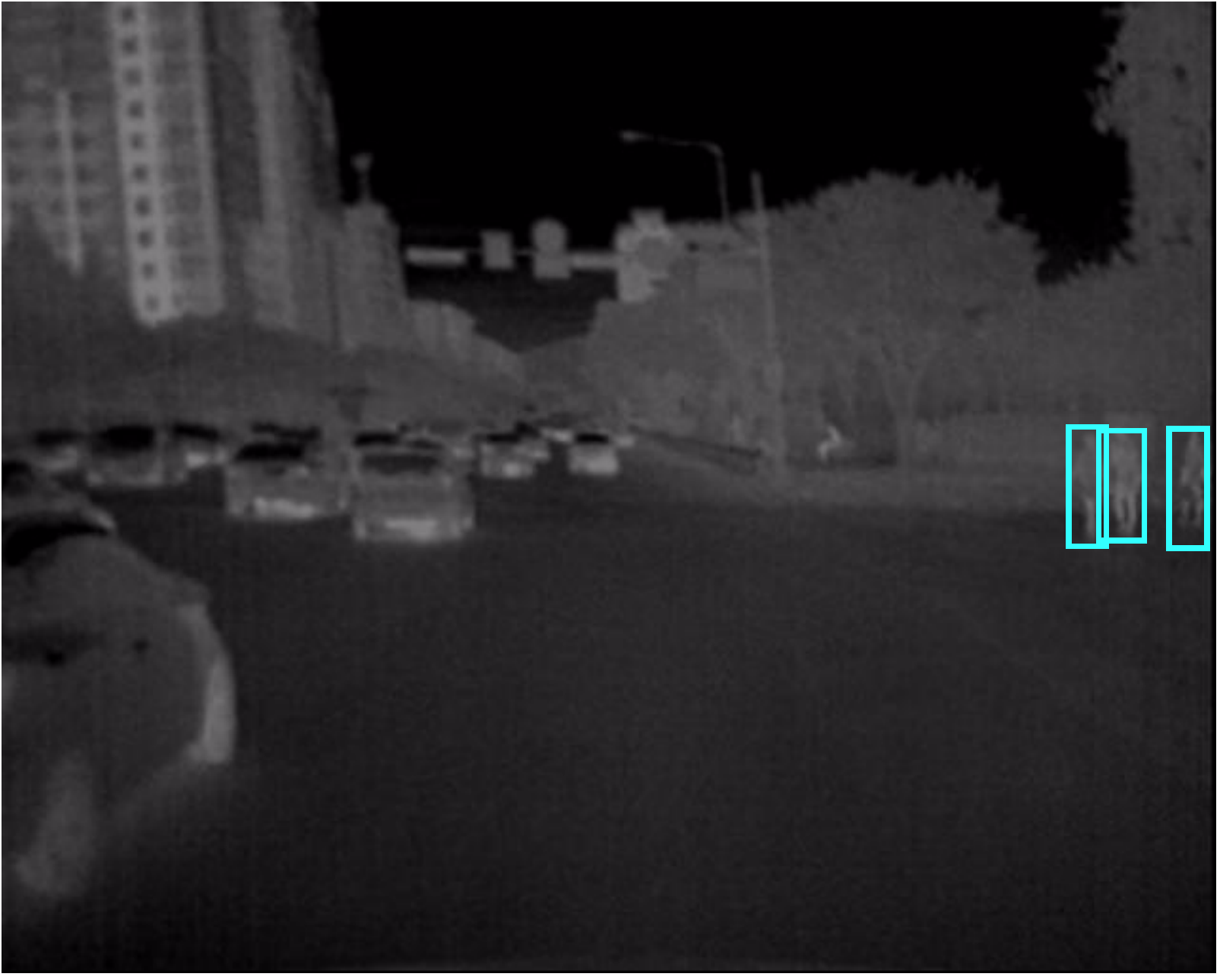}
    \includegraphics[width=0.245\textwidth]{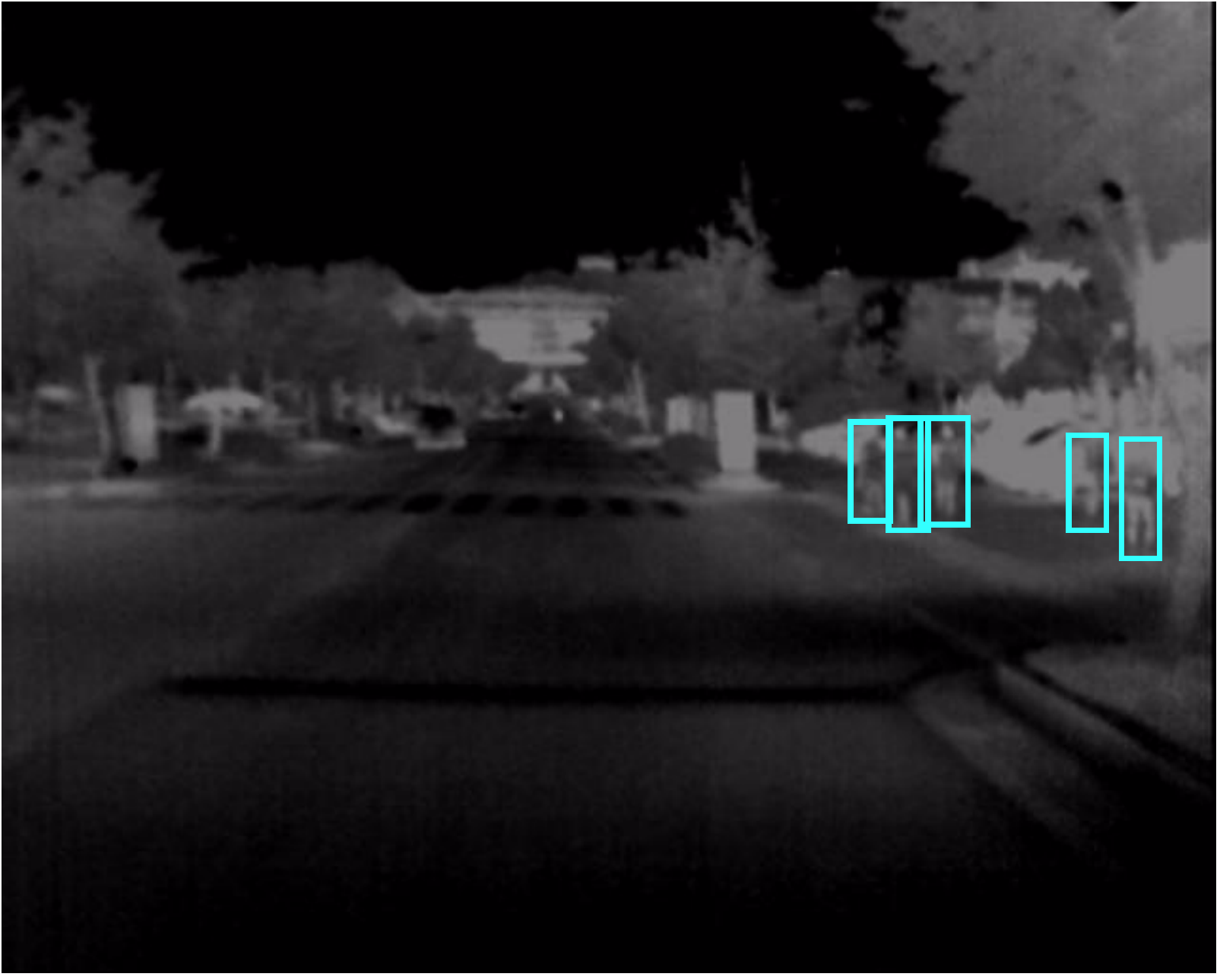}
    \includegraphics[width=0.245\textwidth]{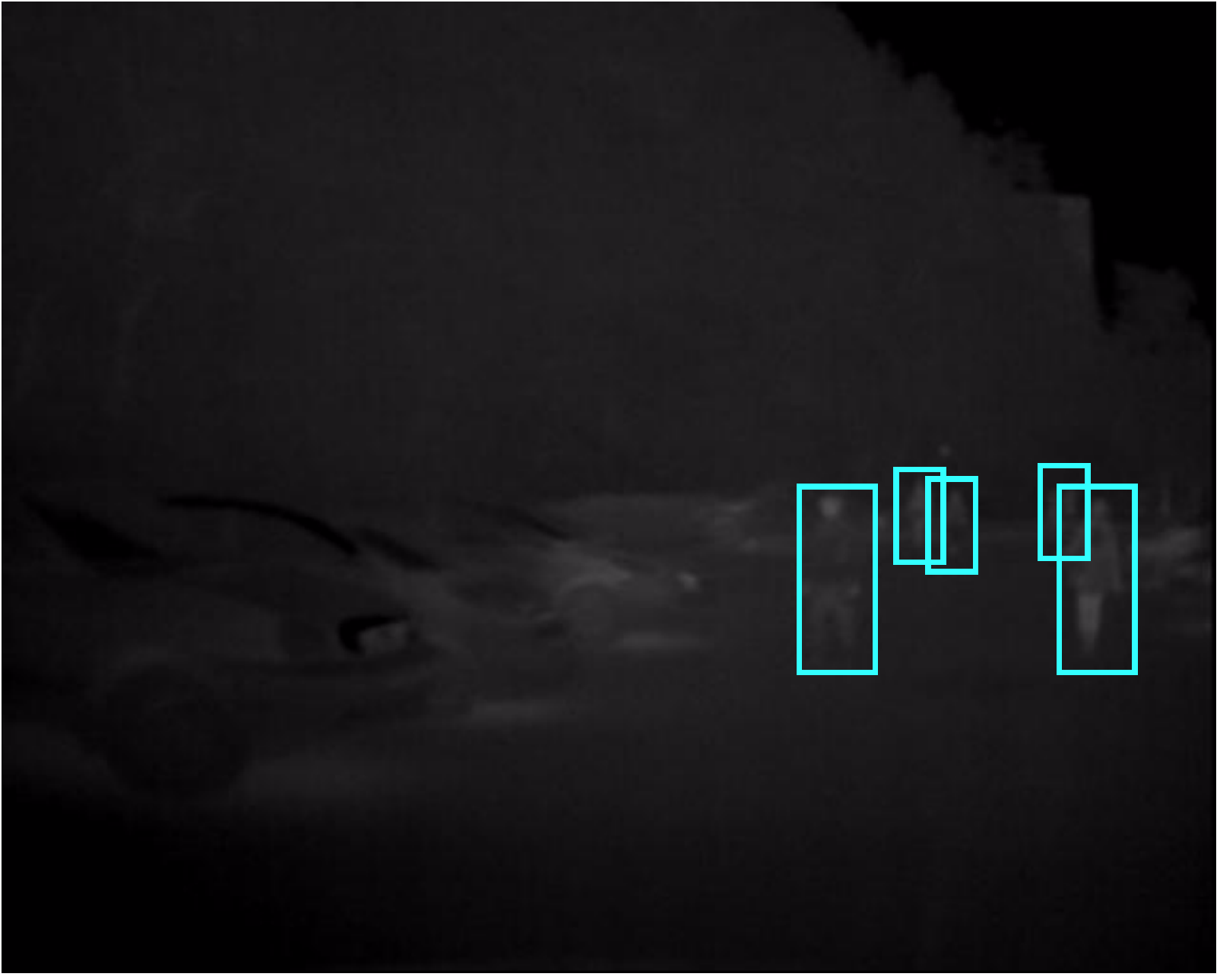}
    \caption{Qualitative results of the proposed method on the KAIST \cite{hwang2015multispectral} dataset.}
    \label{fig:thermal_output_KAIST}
    
  \vspace{-7mm}
    
\end{figure*}

\textbf{KAIST:} Table \ref{tab:detection_multimodality_kaist} compares the proposed method with other state-of-the-art approaches using Miss Rate (MR) metric for the day, night, and all scenarios.  We achieve state-of-the-art results, but the improvement is incremental. Multimodal accuracy is also incremental over unimodal thermal accuracy; in fact, it degrades in night scenarios. As expected, visible unimodal performance is poor at night time and only slightly better at day time. We obtain the best thermal unimodal accuracy compared to other methods; however, we stood third in visible unimodal accuracy. As some of the recent works uses the more commonly used object detection metric, namely  Mean Average Precision (mAP) score, we provide an additional comparative study in Table \ref{tab:detection_multimodality_kaist_map}. In this case, there is a large absolute improvement of $5\%$ in accuracy. 

\begin{table}[t]
\centering

\scalebox{1}{
\begin{tabular}{l|c} 
\bottomrule
\cellcolor{blue!25} $\textbf{Methods}$  & \cellcolor{red!25}$\textbf{Miss Rate}$ $\downarrow$     \\
\toprule

ACF+T+THOG \cite{hwang2015multispectral} & 59.77 \\

Halfway Fusion  \cite{liu2016multispectral_} & 39.80\\ 

MLF-CNN \cite{chen2018multi} & 27.63 \\

\rowcolor{LightGreen}
\textbf{Ours}  & \textbf{25.82} \\
\toprule

\end{tabular}

}
\caption{\textbf{Quantitative comparison of multimodal pedestrian detection  on UTokyo  dataset \cite{takumi2017multispectral}.}}
\label{tab:detection_UTokyo}


\end{table}

\begin{table}[t]
\centering

\scalebox{1}{
\begin{tabular}{l|c|c|c} 
\bottomrule
\cellcolor{blue!25} $\textbf{Methods}$  & \cellcolor{red!25}$\textbf{MR All}$ $\downarrow$  & \cellcolor{red!25}$\textbf{MR Day}$ $\downarrow$ & \cellcolor{red!25}$\textbf{MR Night}$ $\downarrow$   \\
\toprule

MACF \cite{park2018unified} & 60.1 & 61.3 & 48.2\\

Choi et al.  \cite{choi2016multi} & 47.3  & 49.3 & 43.8 \\ 

Halfway Fusion  \cite{park2018unified} & 37.0  & 38.1 & 34.4 \\ 

Park et al.  \cite{park2018unified} & 31.4  & 31.8 & 30.8 \\ 
AR-CNN \cite{zhang2019weakly_} & 22.1 & 24.7 & 18.1 \\

MBNet \cite{zhou2020improving} & 21.1 & 24.7 & 13.5 \\

\rowcolor{LightGreen}
\textbf{Ours}  & \textbf{19.04} & \textbf{20.32} & \textbf{12.86}\\
\toprule

\end{tabular}
}
\caption{\textbf{
Quantitative comparison of pedestrian detection  on CVC-14  dataset \cite{gonzalez2016pedestrian}. 
}
MR is Miss Rate.}

\vspace{4mm}

\label{tab:detection_cvc14}

\end{table}

\textbf{UTokyo:} Table \ref{tab:detection_UTokyo} compares the proposed method with other state-of-the-art multimodal approaches using the Miss Rate metric. The proposed method achieves an absolute improvement of roughly $2\%$ relative to the previous state-of-the-art method \cite{chen2018multi}.

\textbf{CVC-14:} Table \ref{tab:detection_cvc14} compares the proposed method with other state-of-the-art multimodal approaches using the Miss Rate metric for the day, night, and all scenarios. It is to be noted that despite the alignment issue between color and thermal images due to improper camera calibration, the proposed approach outperformed other multimodal pedestrian detectors without performing any pre-processing or post-processing. An absolute improvement of $2\%$ was obtained over the previous state-of-the-art method \cite{zhou2020improving}.


\begin{figure}
    \centering
    
    \includegraphics[width=0.24\textwidth]{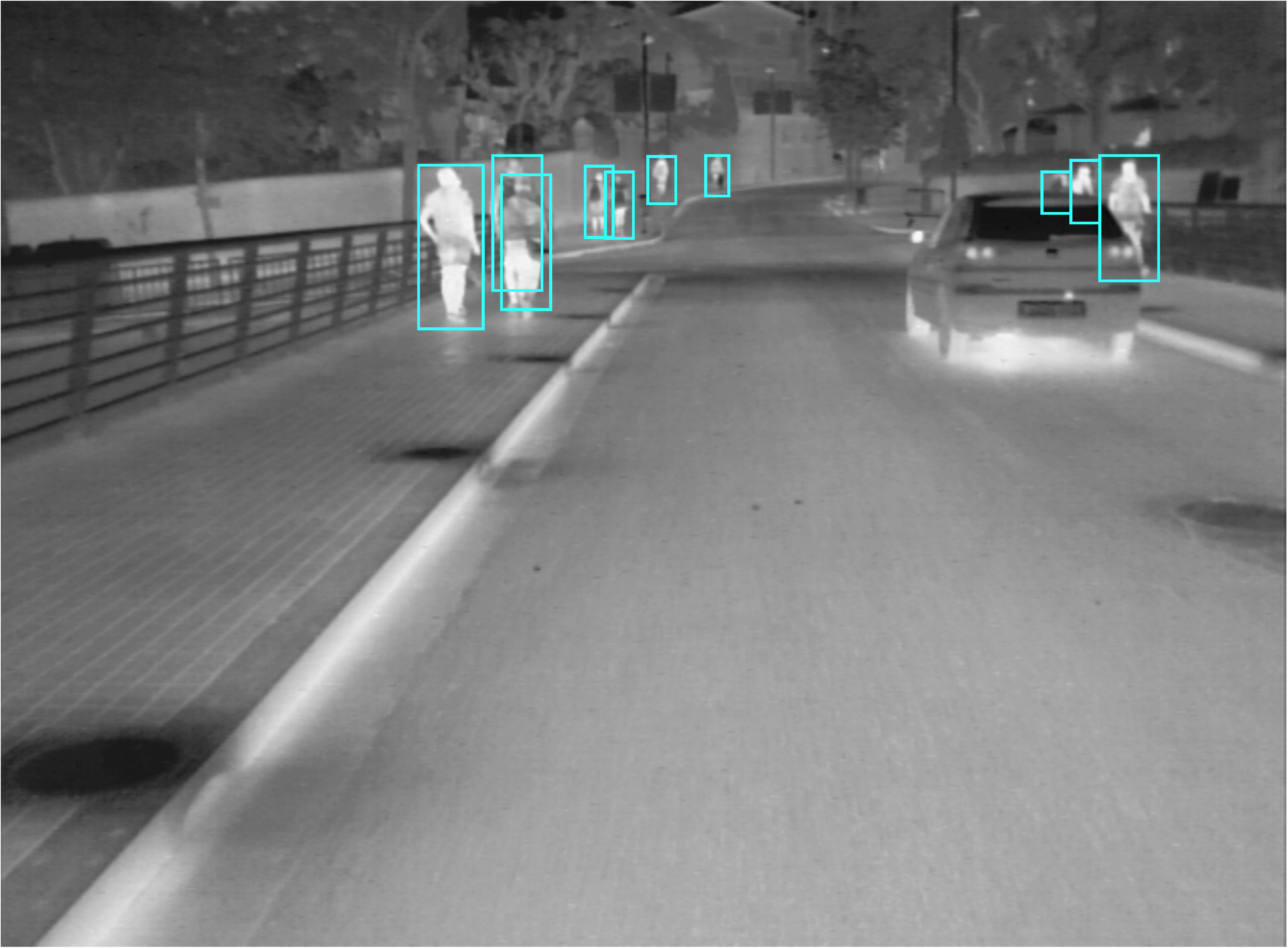}
    \includegraphics[width=0.24\textwidth]{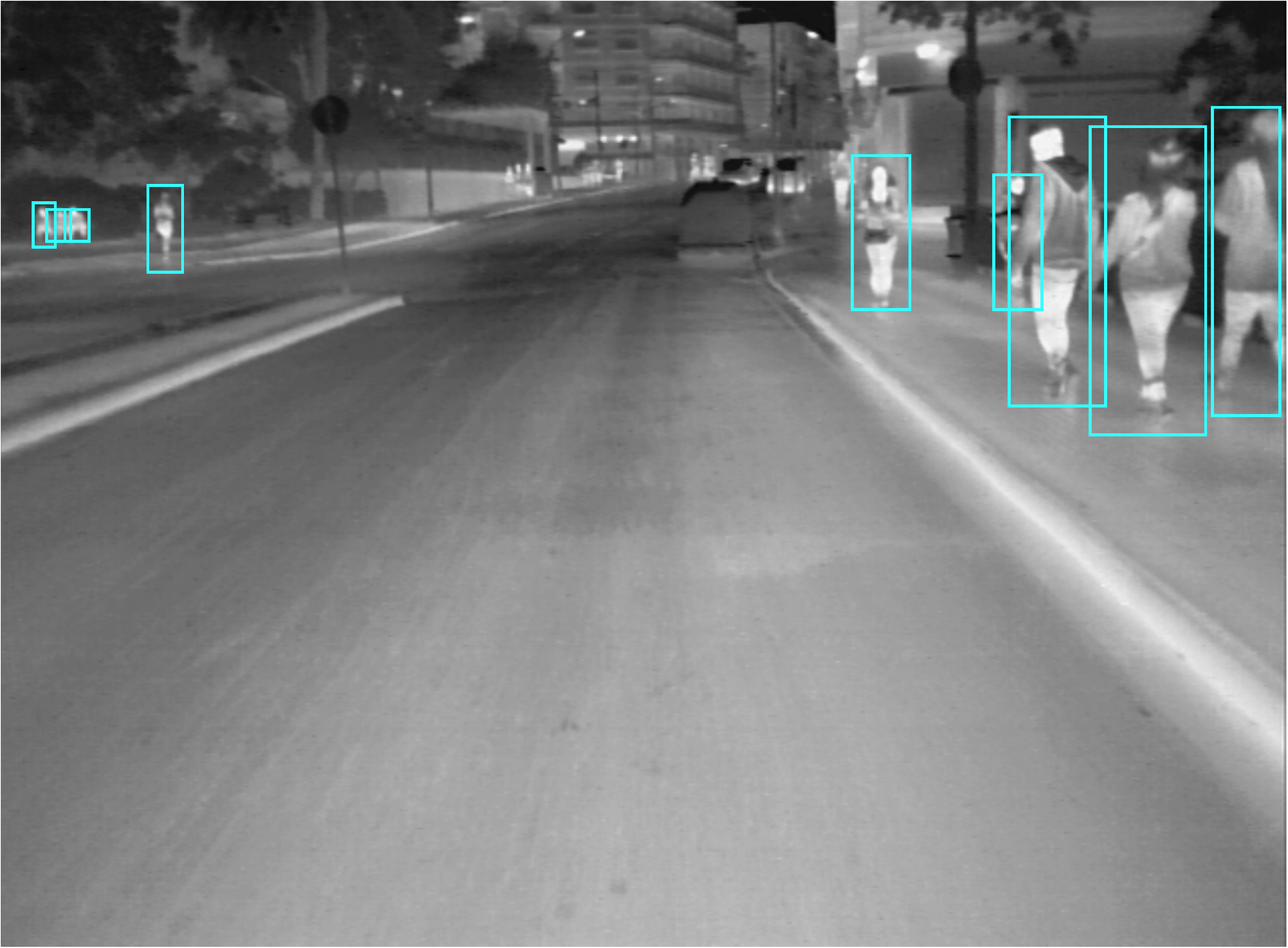}
    
     \vspace{0.25 mm}
    
    \includegraphics[width=0.24\textwidth]{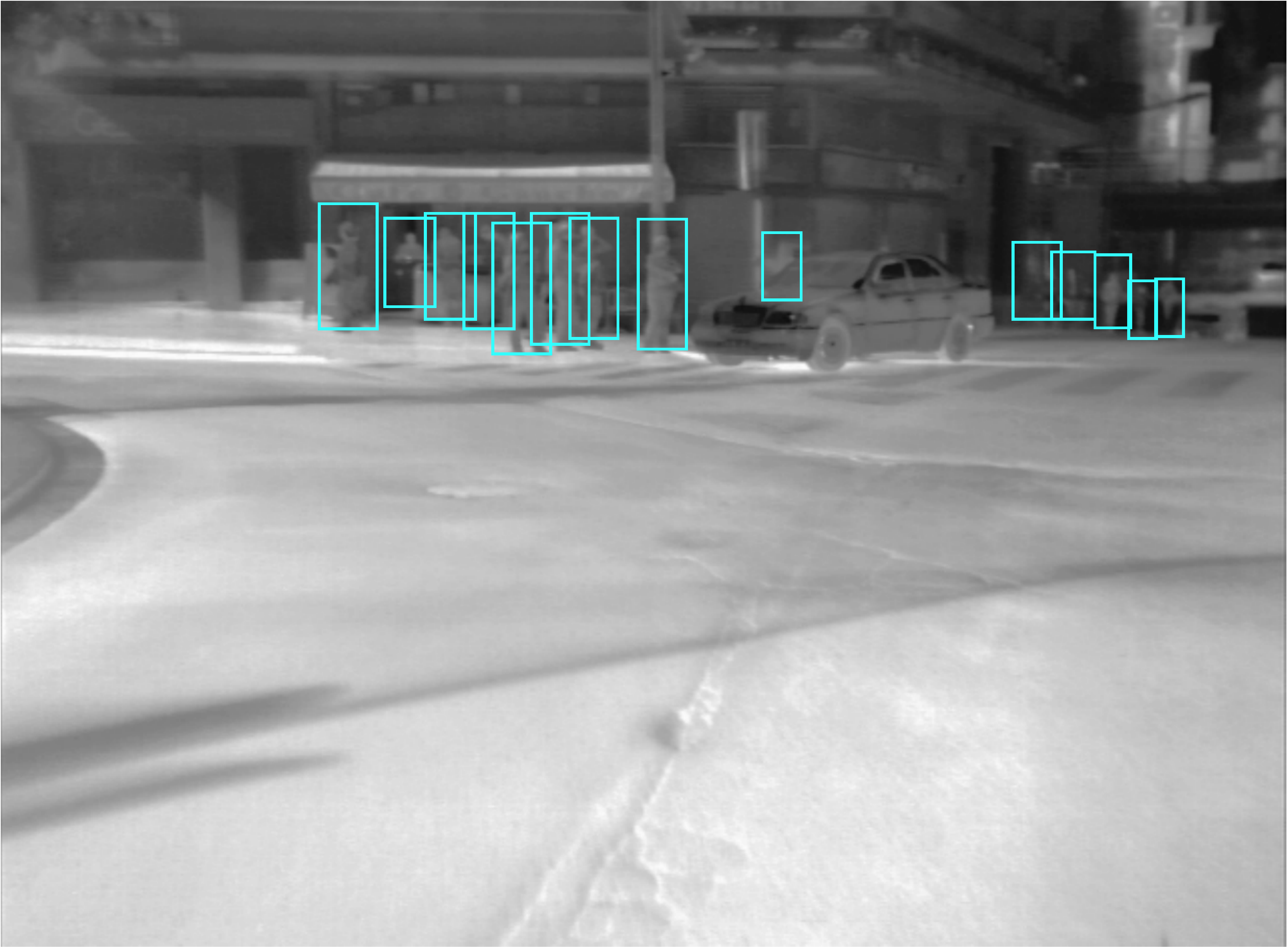}
    \includegraphics[width=0.24\textwidth]{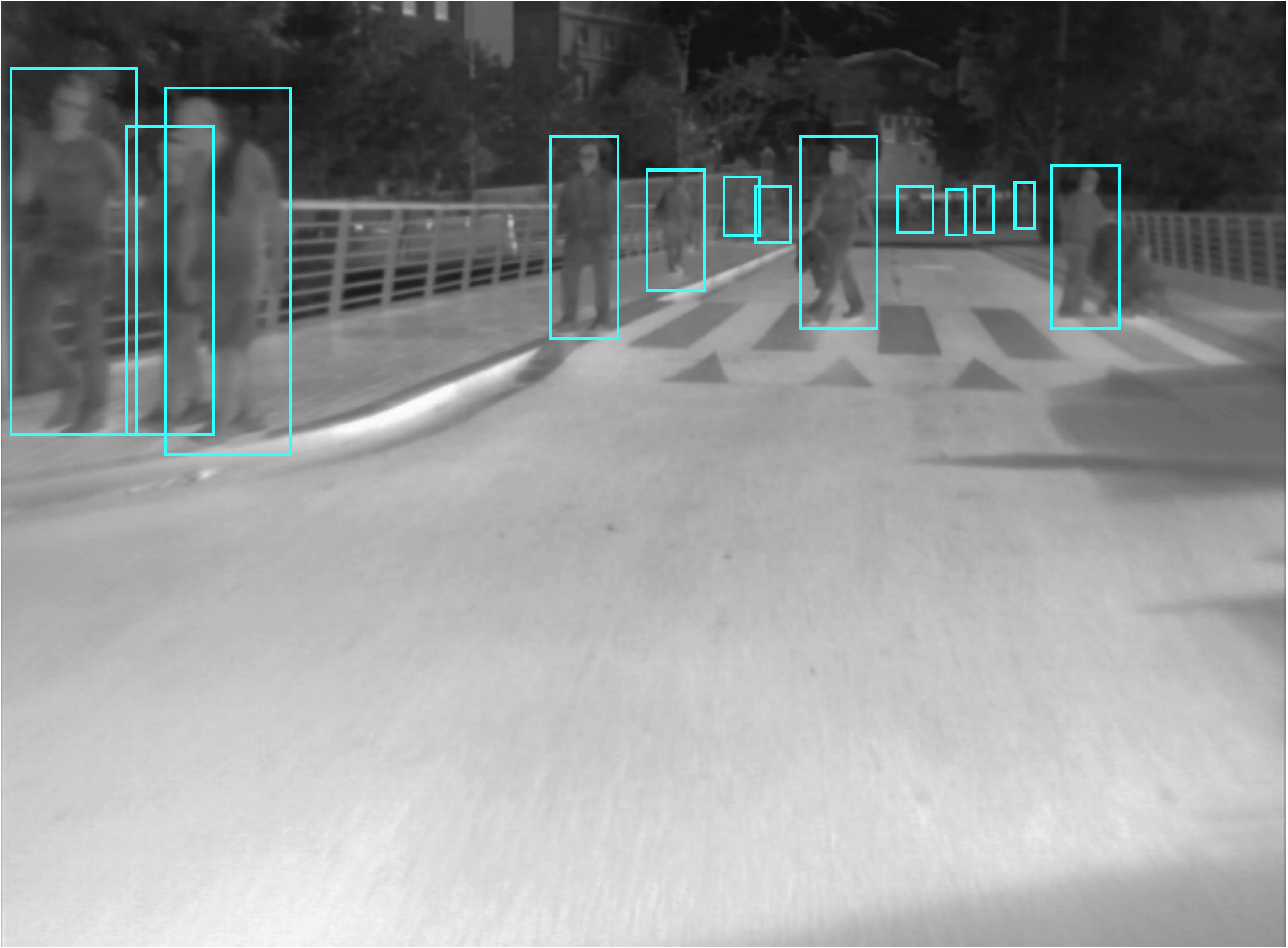}
    
    \caption{
    Qualitative results on the CVC-14 \cite{gonzalez2016pedestrian} dataset. 
    }
    
 \vspace{4 mm}

    \label{fig:thermal_output_CVC-14}
\end{figure}


    
    
    
    

    
    
        


\subsection{Qualitative study}

\begin{figure}
    \centering
    
 \includegraphics[width=0.24\textwidth]{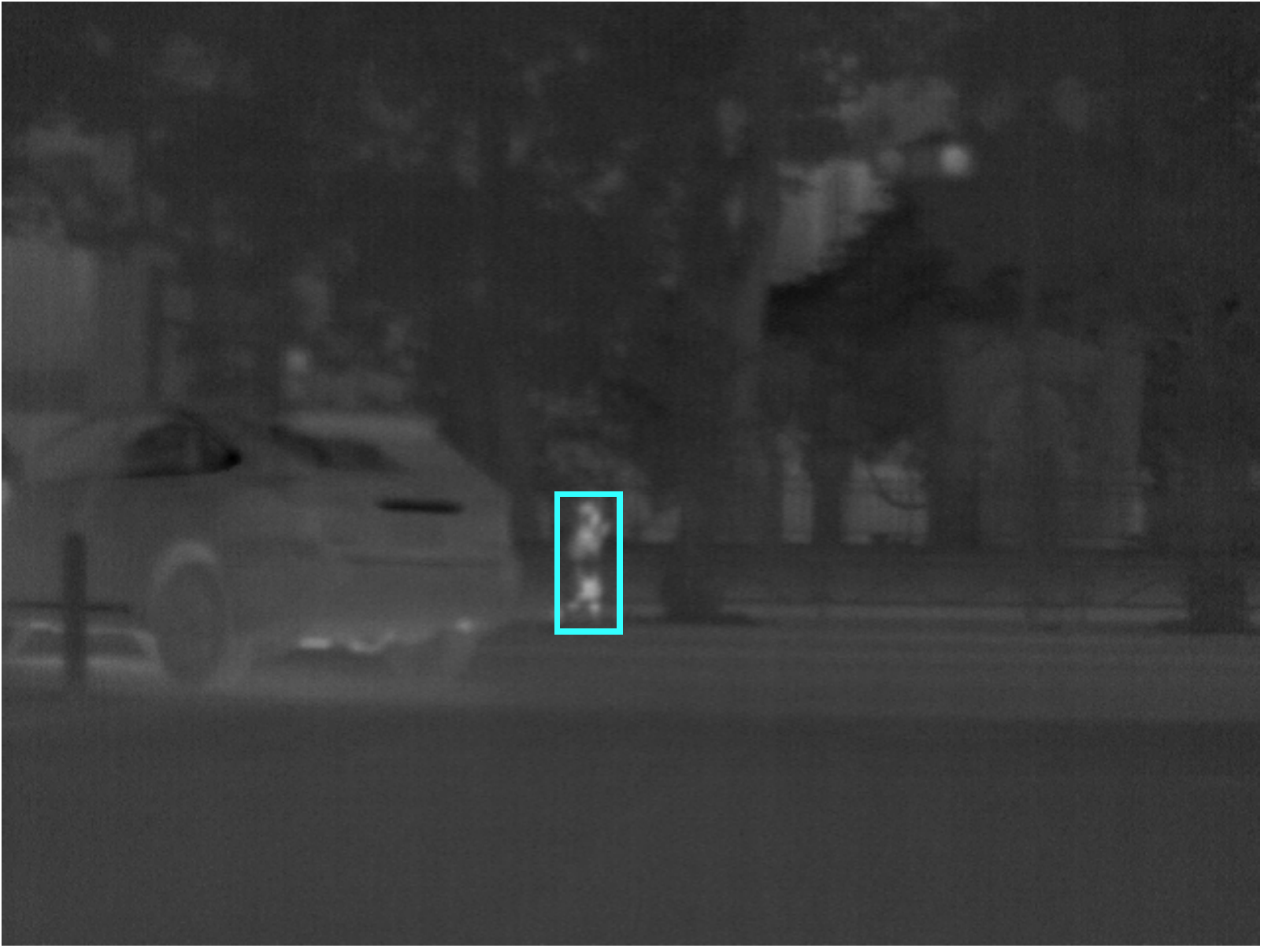}
    \includegraphics[width=0.24\textwidth]{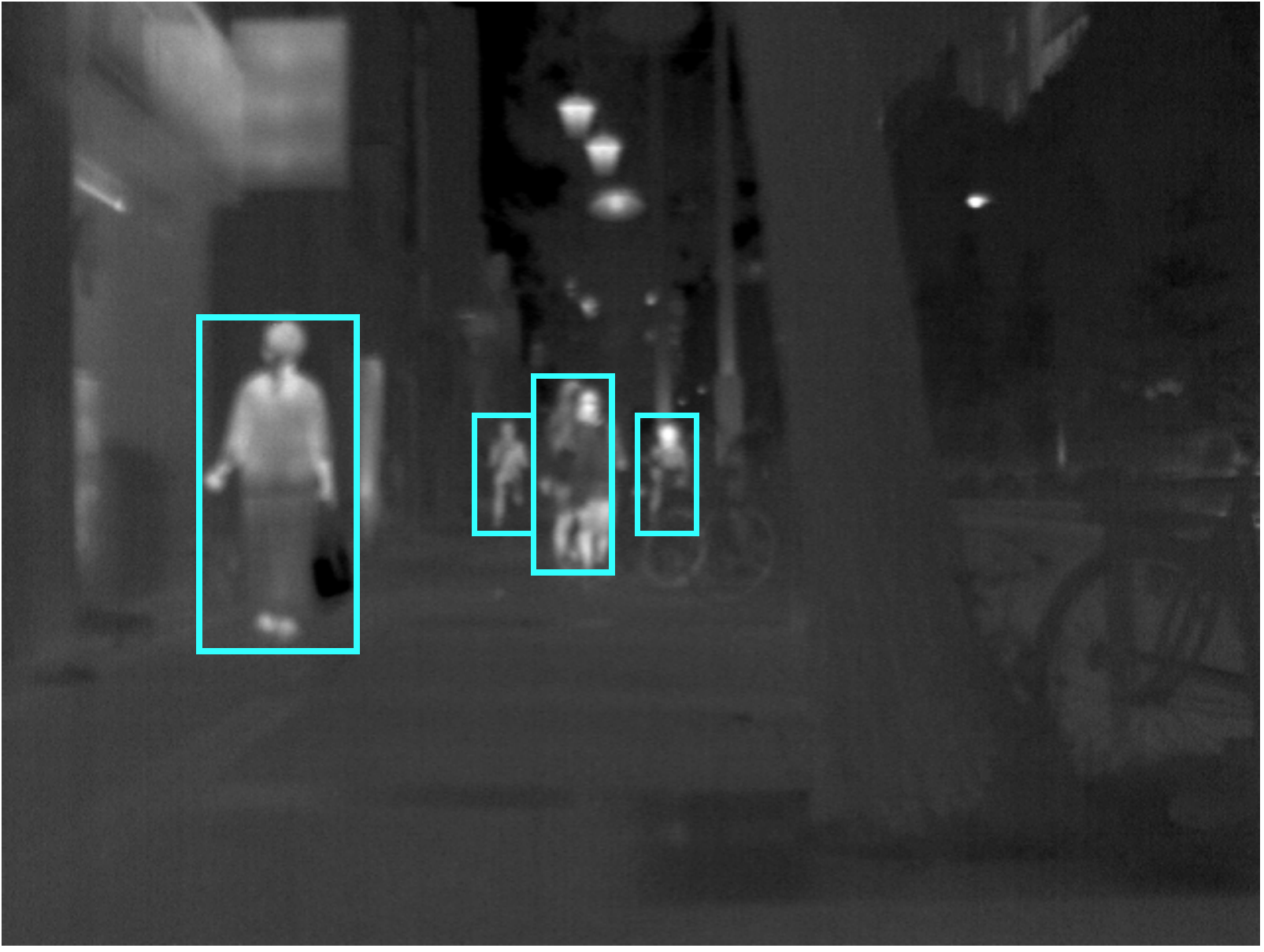}
    
     \vspace{0.25 mm}
        
    \includegraphics[width=0.24\textwidth]{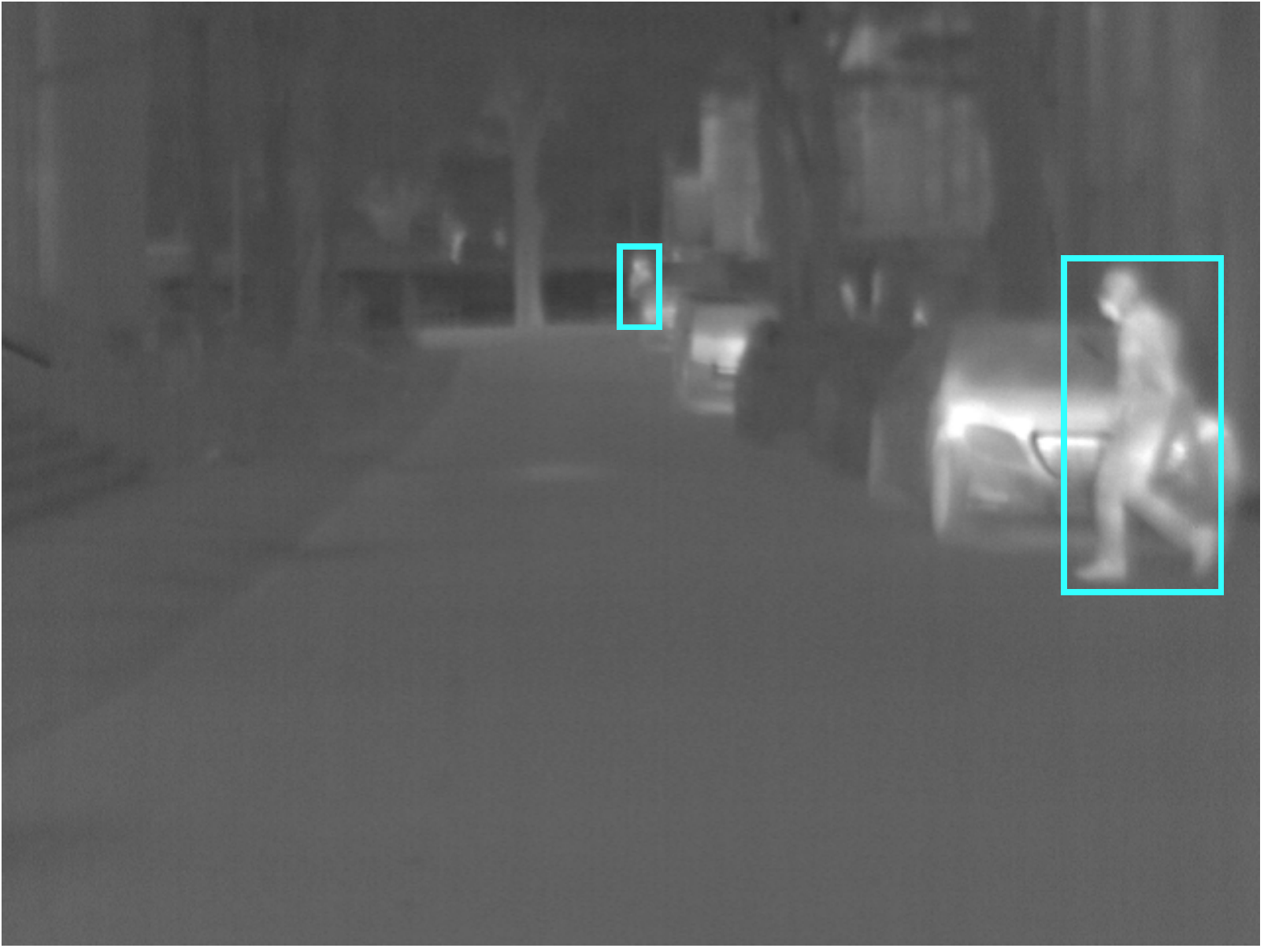}
    \includegraphics[width=0.24\textwidth]{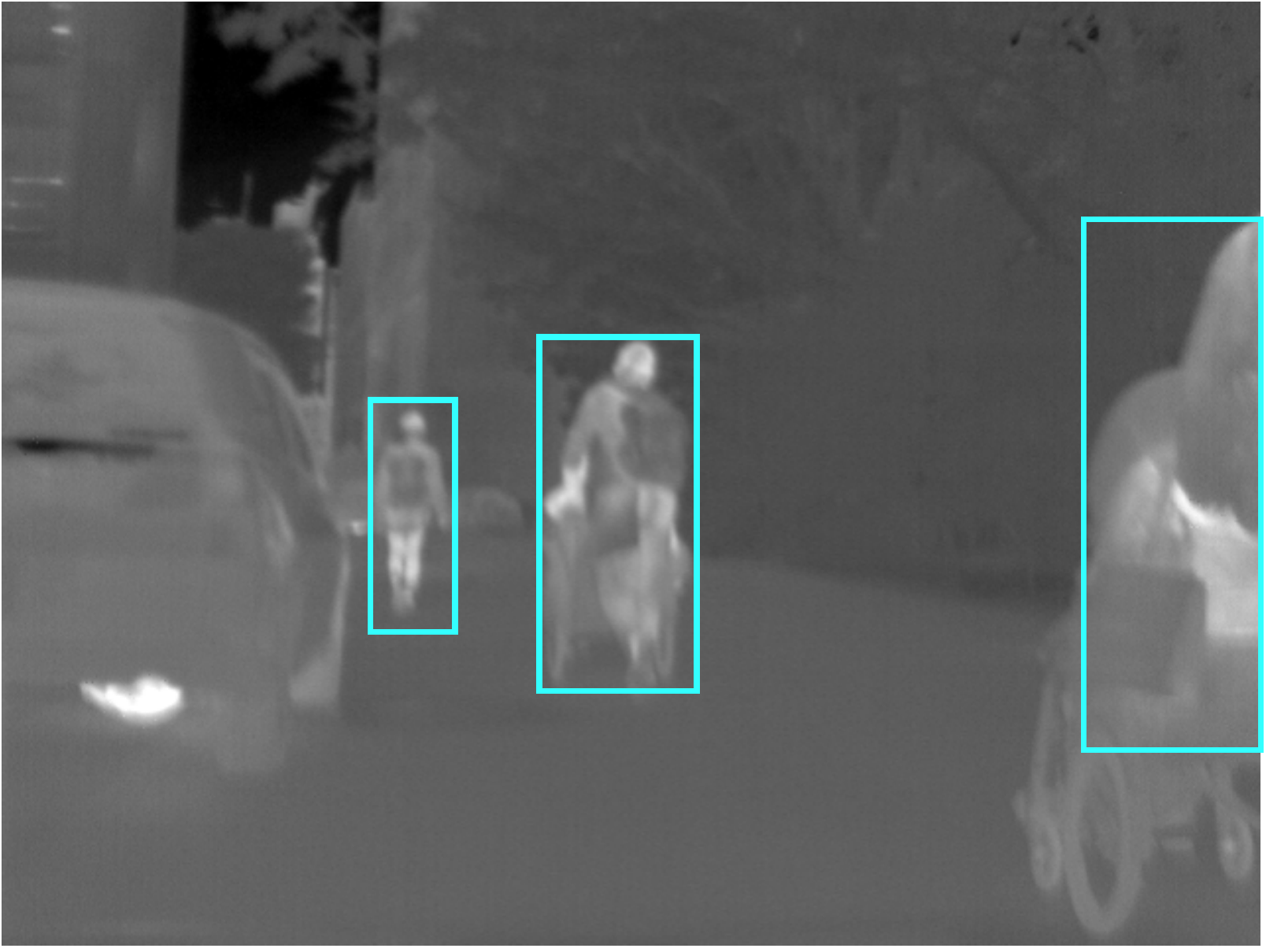}

    \caption{Qualitative results  on the UTokyo \cite{takumi2017multispectral} dataset.}
    \label{fig:thermal_output}
    
\vspace{3mm}

\end{figure}

Figures \ref{fig:thermal_output_KAIST}, \ref{fig:thermal_output_CVC-14}  and \ref{fig:thermal_output}    shows qualitative results of the proposed model on KAIST, CVC-14 and UTokyo dataset respectively. More qualitative results are provided in {\url{https://youtu.be/FDJdSifuuCs}}. Top and bottom rows of Figures \ref{fig:thermal_output_CVC-14} and \ref{fig:thermal_output} shows night and daytime scenes respectively. The proposed method provides accurate detection of pedestrians even in challenging scenarios. Small pedestrians are one such challenge, and good detection can be observed in Figure \ref{fig:thermal_output} where a small child (top left) and a far pedestrian (bottom left) are detected accurately. Another challenging scenario is a crowded group of pedestrians with occlusions, and we can observe good detections in Figure \ref{fig:thermal_output_CVC-14}. \textit{SCoFA} module aids in learning context to provide robust detection in these scenarios. However, we occasionally observe a few missed detections in groups of pedestrians. In some cases, a group of pedestrians is detected as one pedestrian. Finally, partially visible or occluded pedestrians are detected well when there is sufficient visibility, as observed in the bottom right image of Figure \ref{fig:thermal_output}.

\begin{figure}
    \captionsetup{singlelinecheck=false, font=small}
    \centering
    \includegraphics[width=1.14in, height=1.1in]{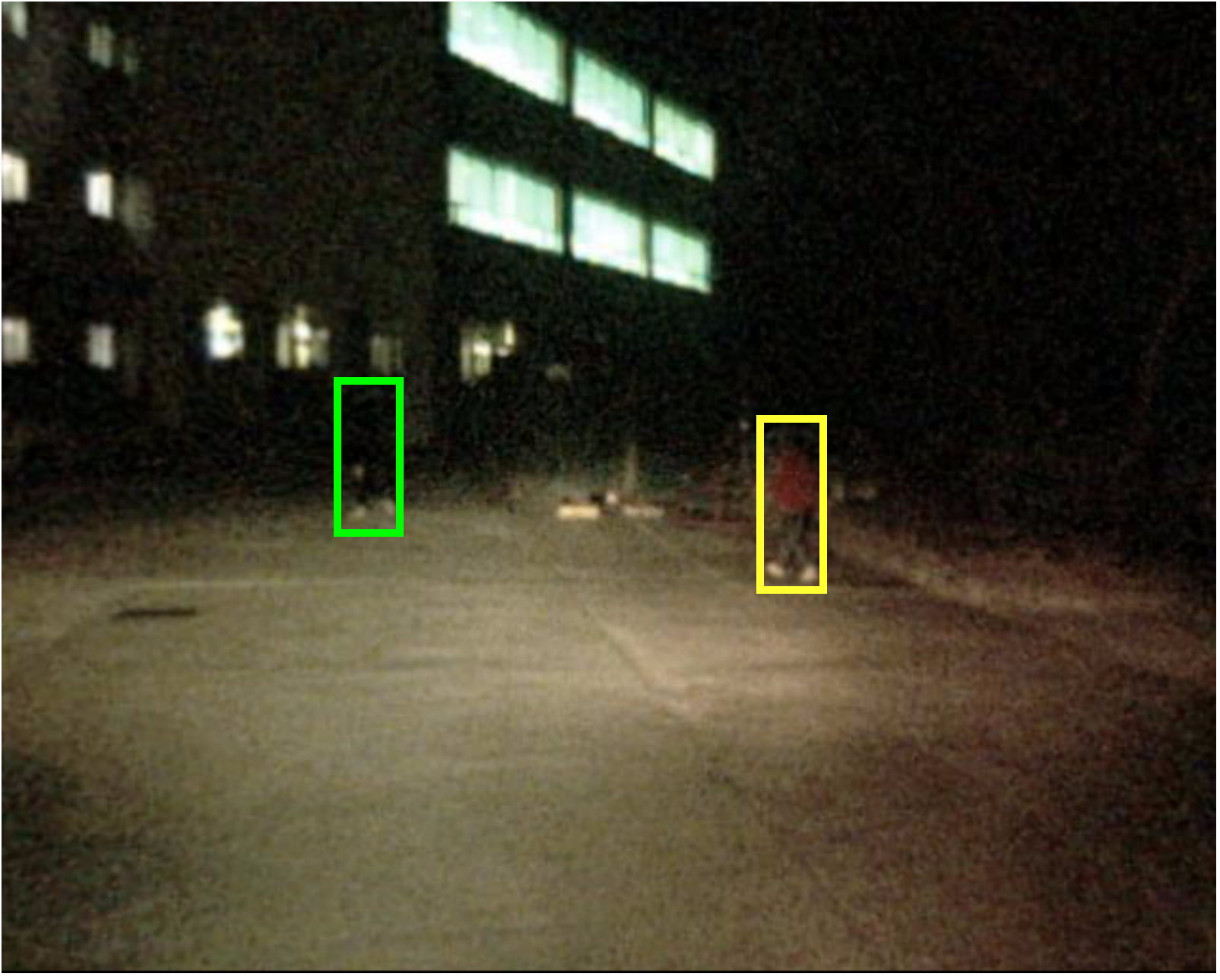}
    \hspace{-0.078in}
    \includegraphics[width=1.14in, height=1.1in]{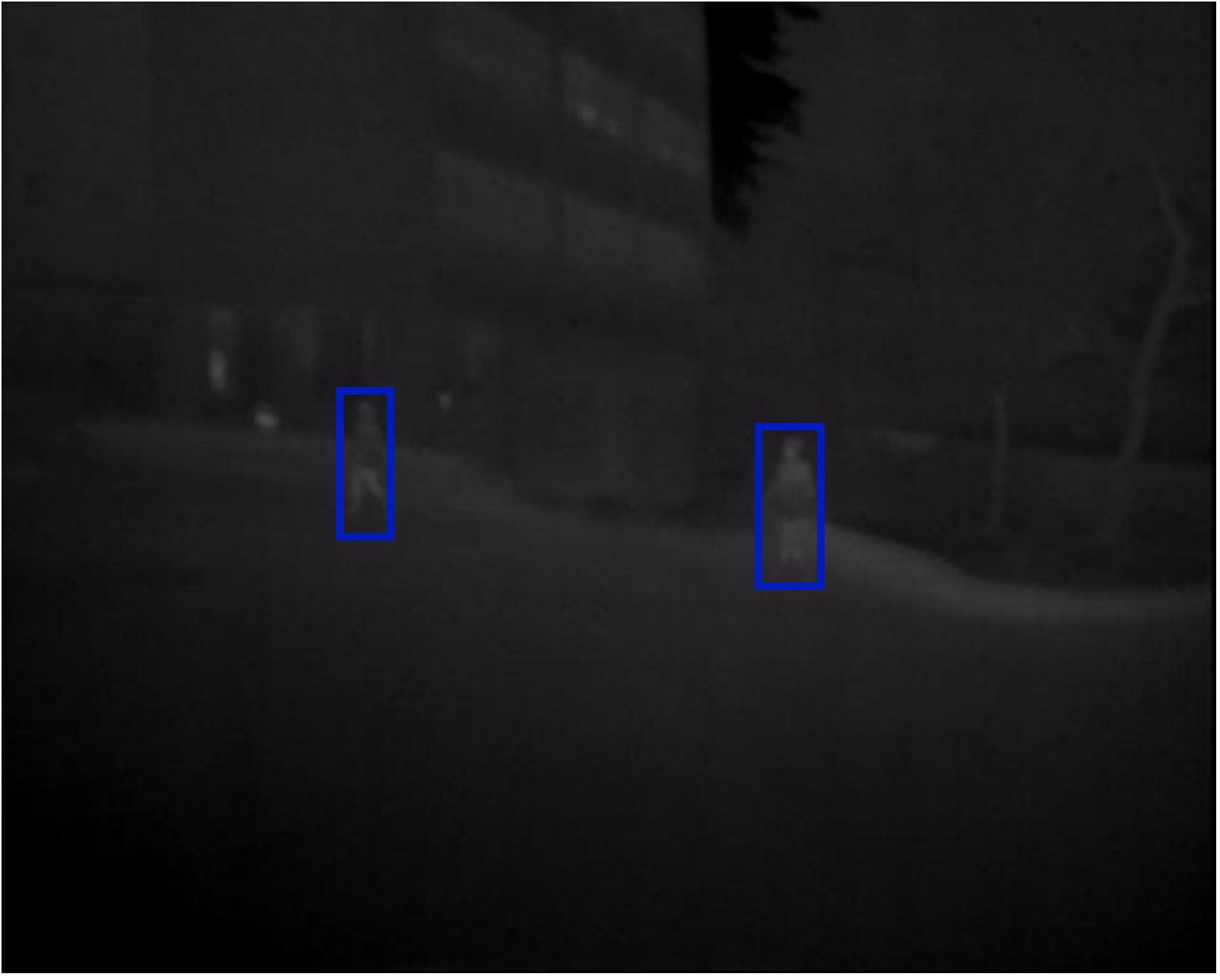}
    \hspace{-0.078in}
    \includegraphics[width=1.14in, height=1.1in]{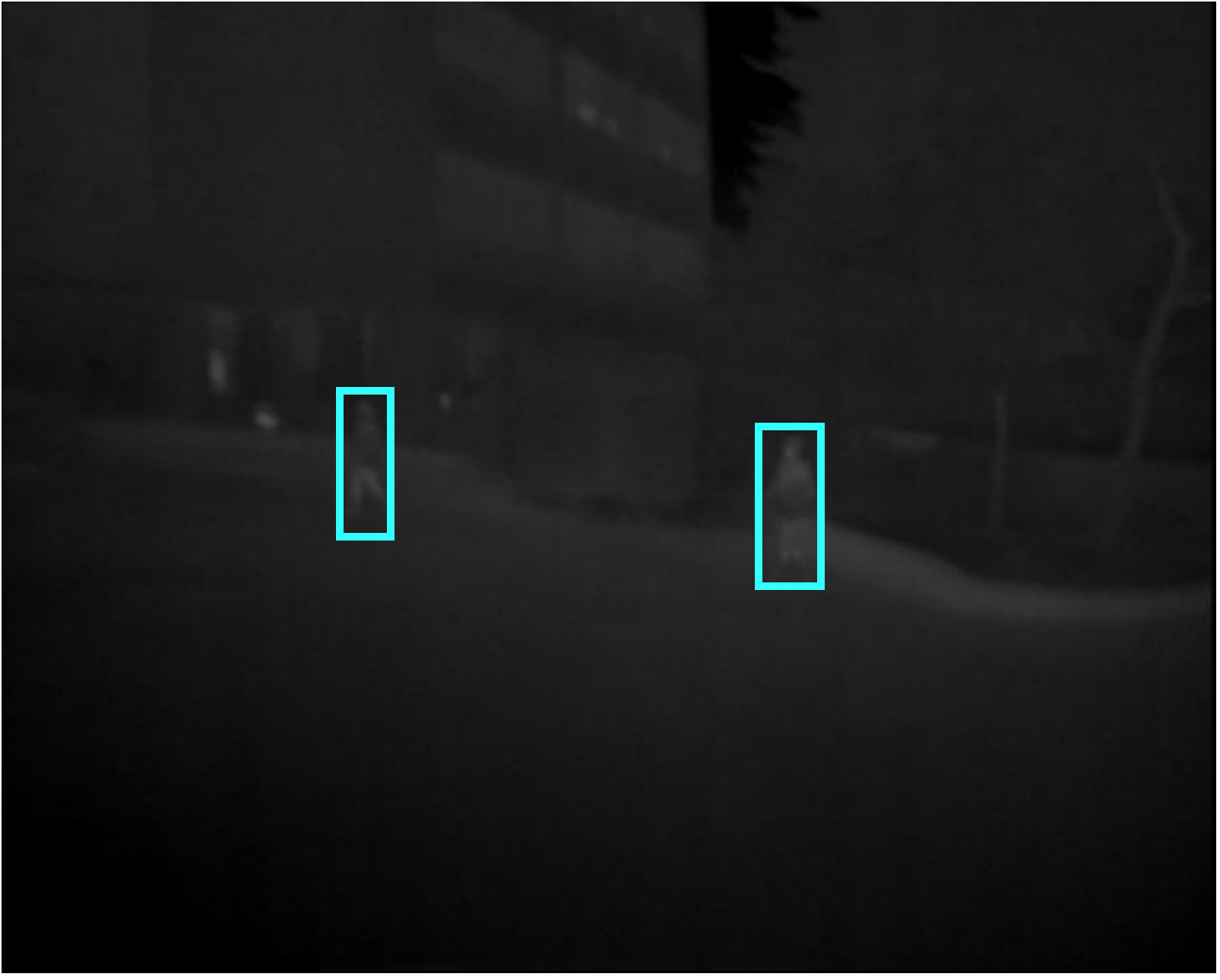}\\
    \vspace{0.02in}
      \includegraphics[width=1.14in, height=1.1in]{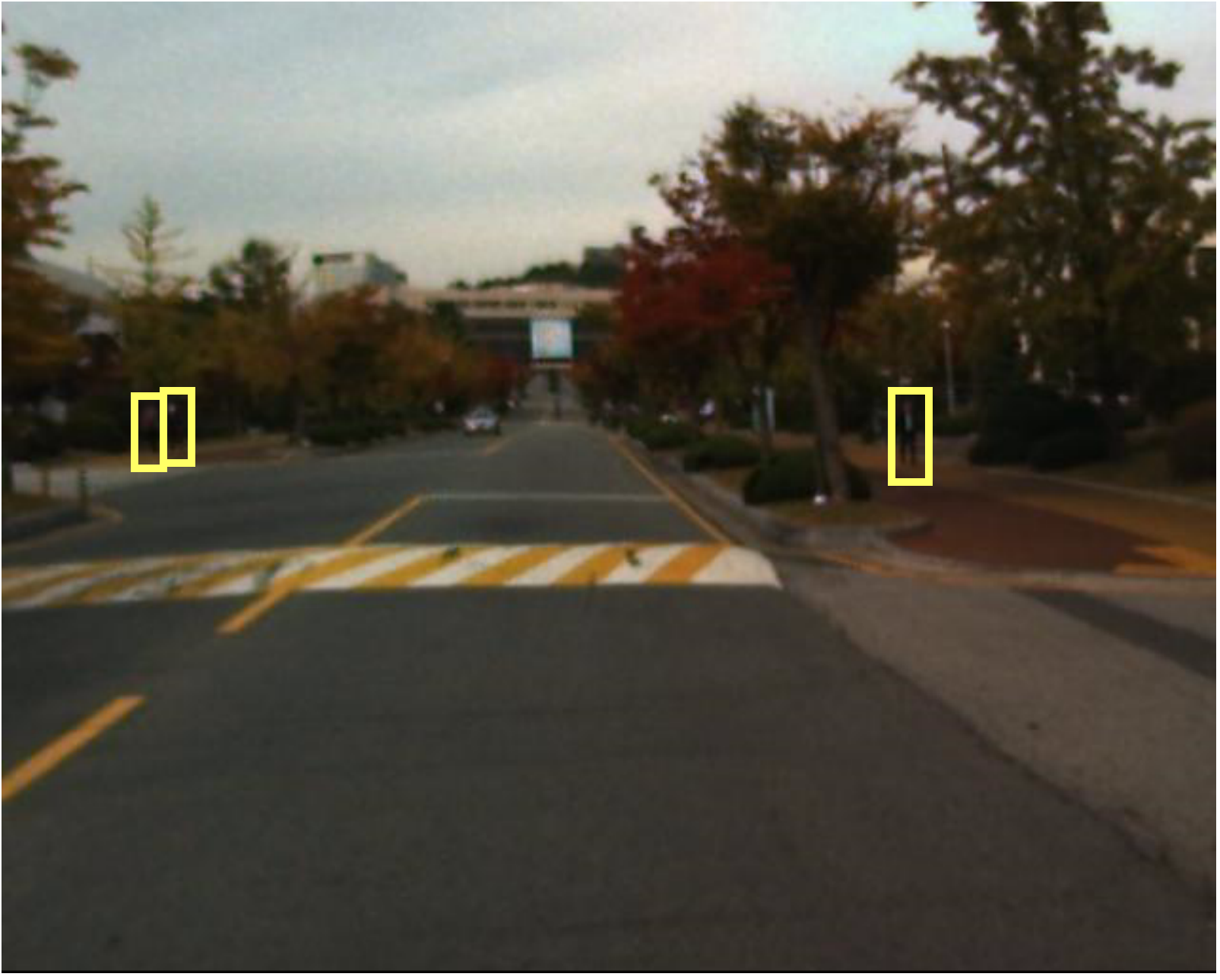}
    \hspace{-0.078in}
    \includegraphics[width=1.14in, height=1.1in]{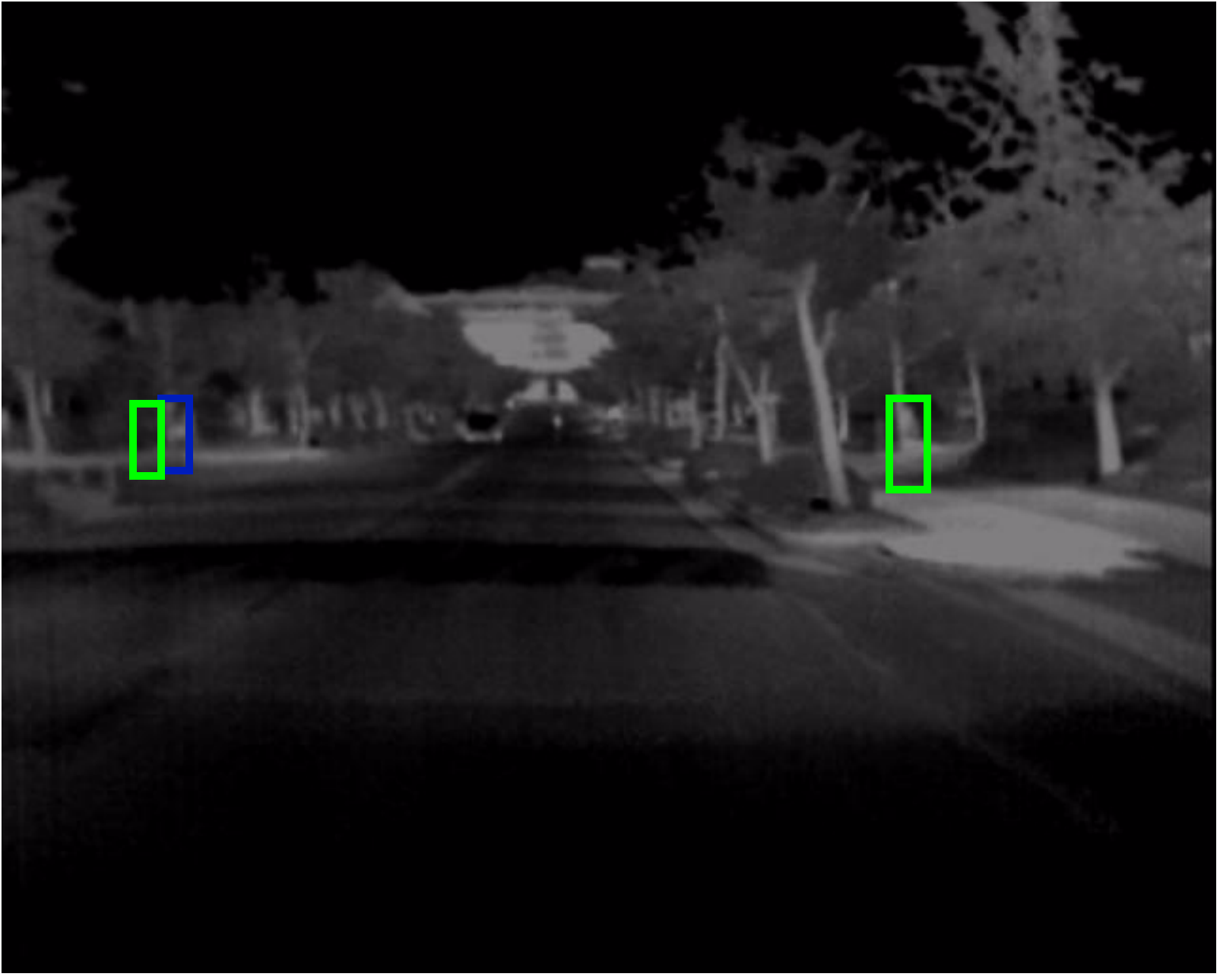}
    \hspace{-0.078in}
    \includegraphics[width=1.14in, height=1.1in]{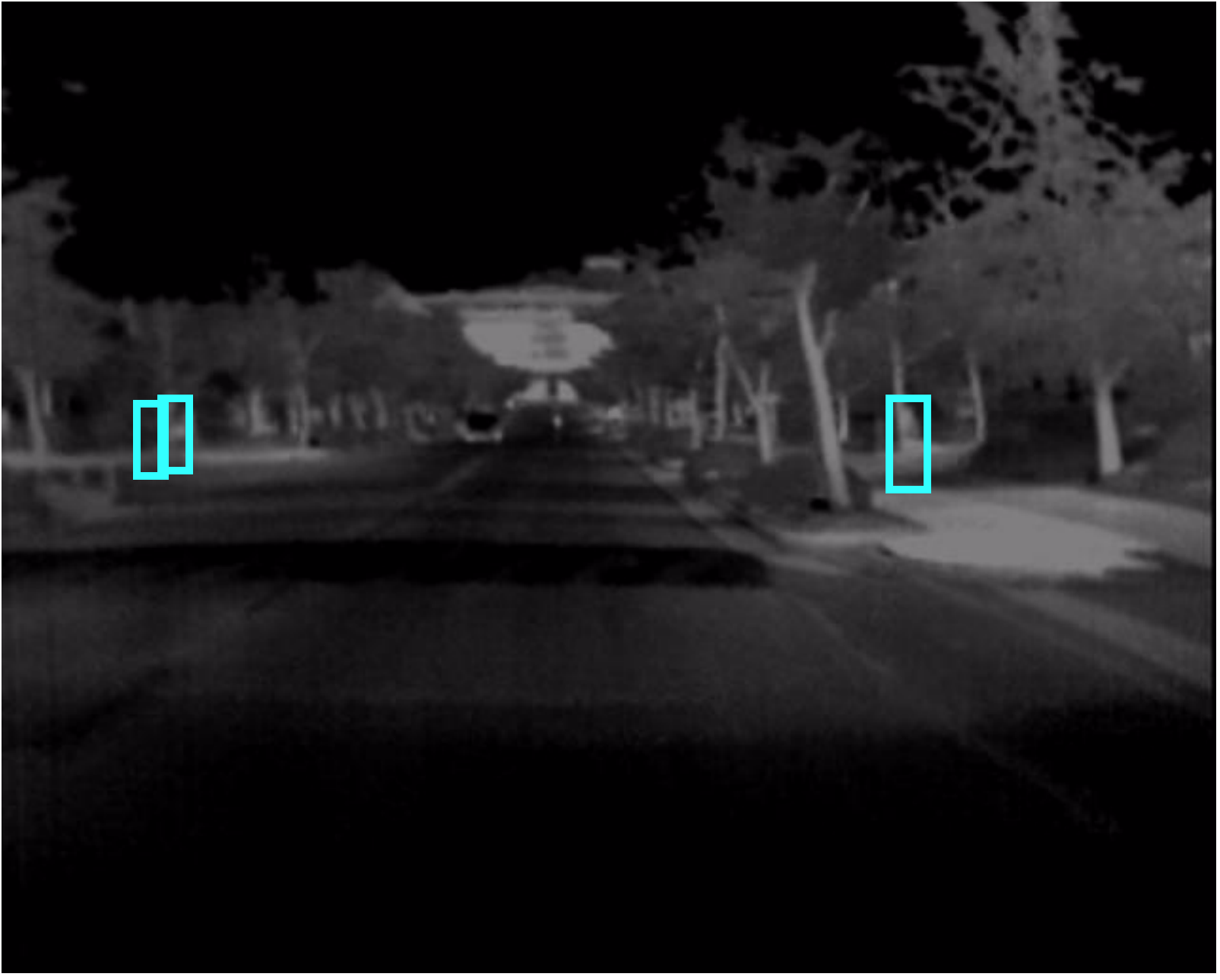}\\
    \vspace{0.02in}
      \includegraphics[width=1.14in, height=1.1in]{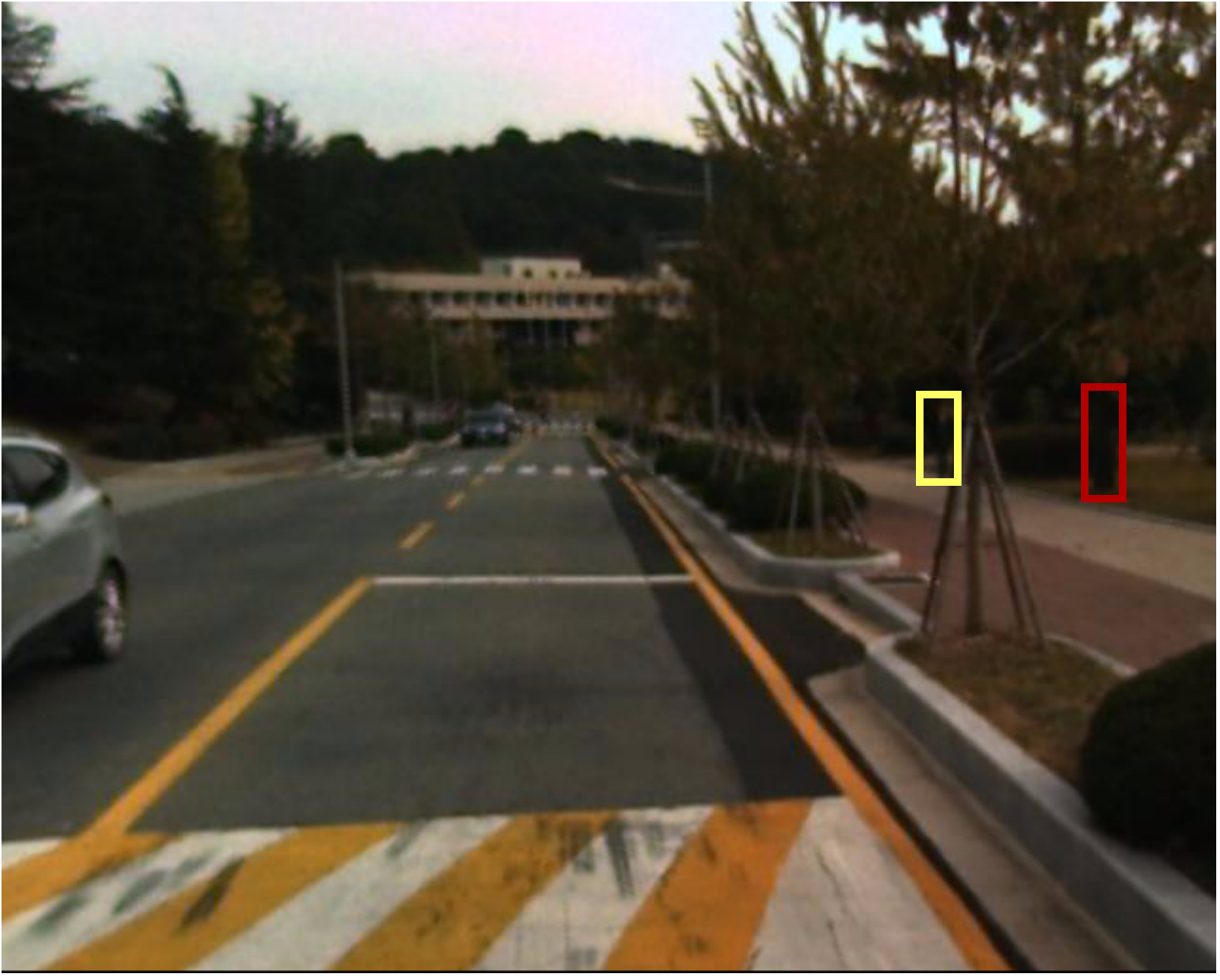}
    \hspace{-0.078in}
    \includegraphics[width=1.14in, height=1.1in]{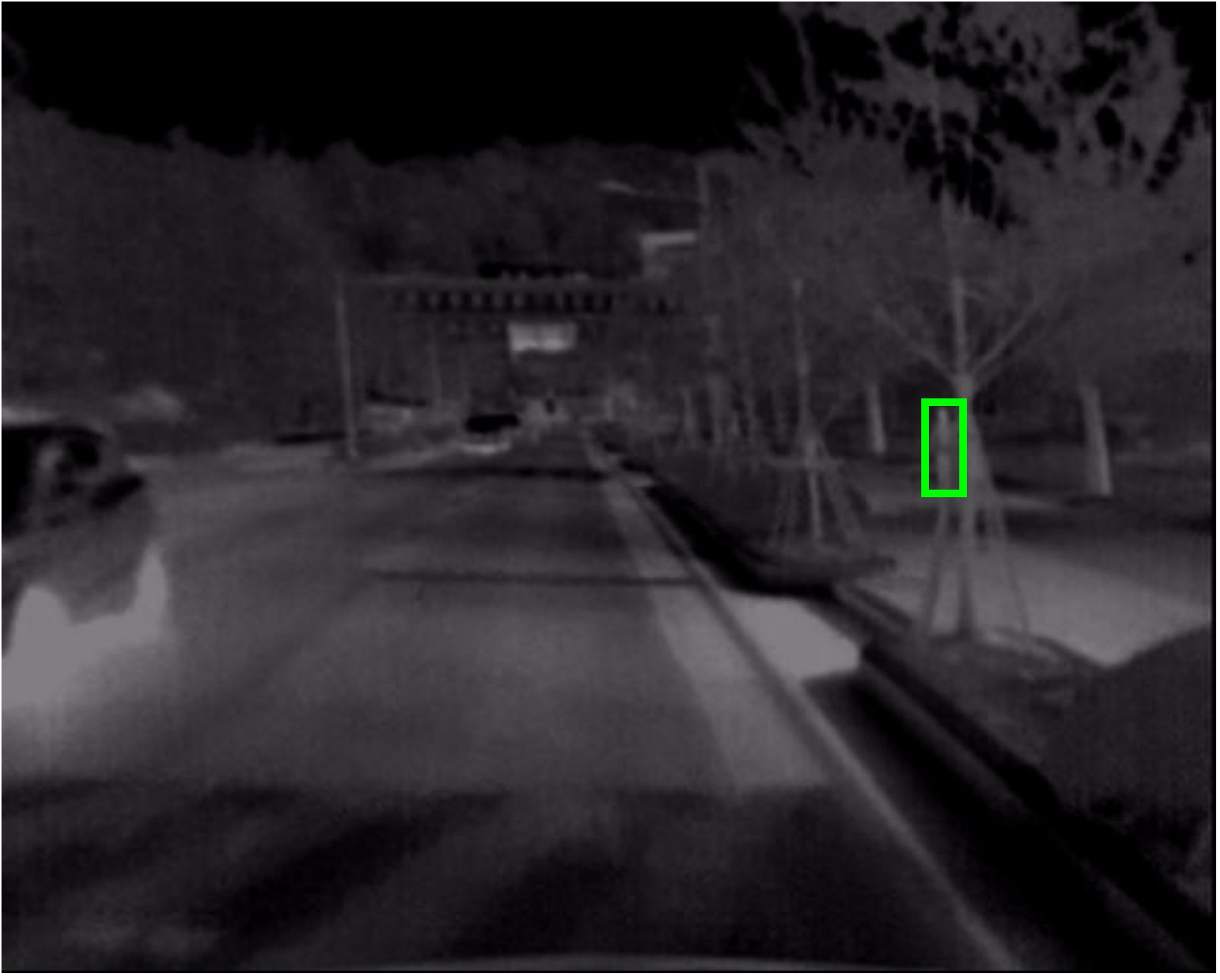}
    \hspace{-0.078in}
    \includegraphics[width=1.14in, height=1.1in]{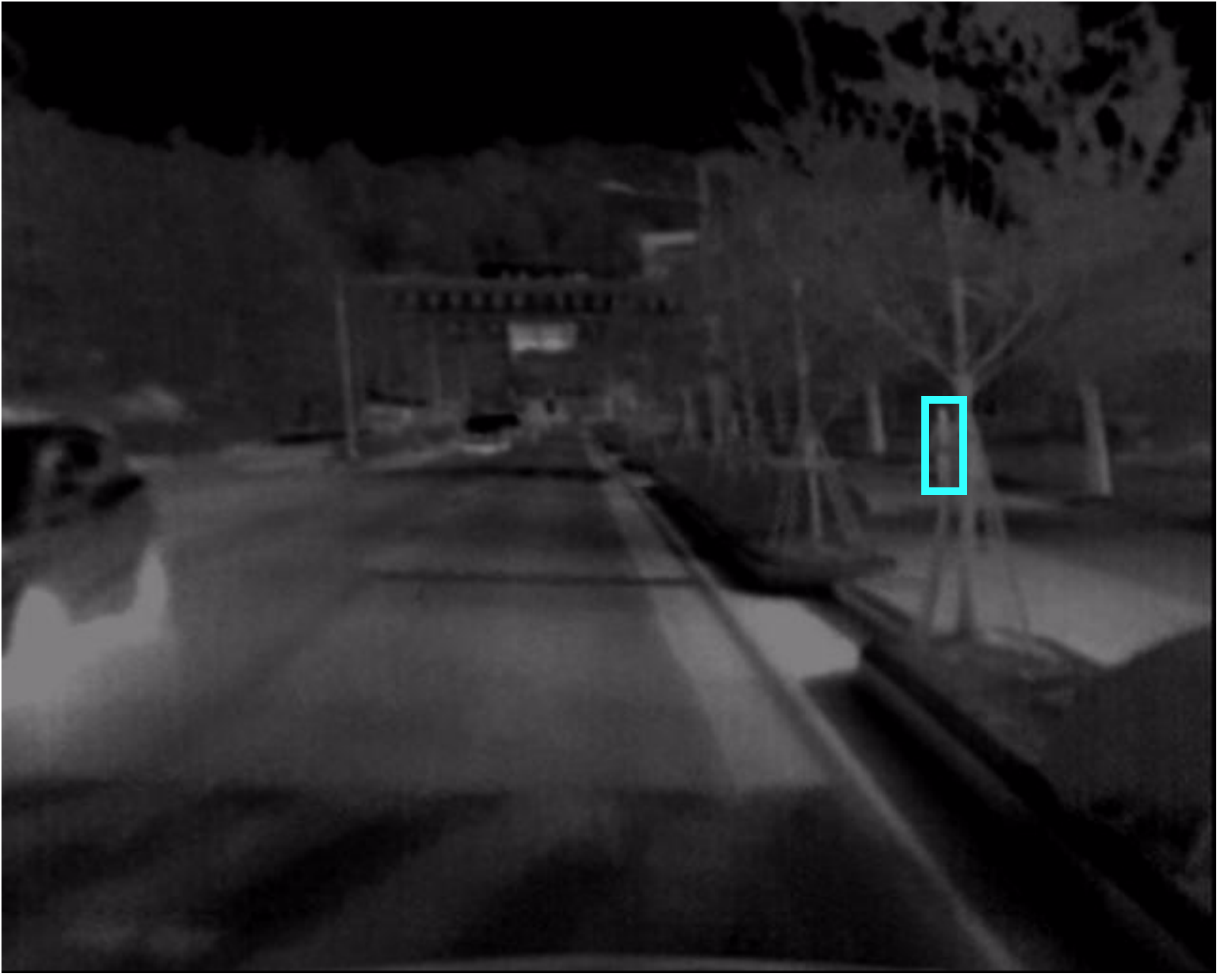}\\
    \caption{Comparison of unimodal and multimodal detections on KAIST dataset \cite{hwang2015multispectral}. \textcolor{yellow}{Yellow}, \textcolor{blue}{blue} \& \textcolor{cyan}{cyan} colors are for the correct detections from RGB, thermal and both inputs respectively. \textcolor{green}{Green} \& \textcolor{red}{red} bounding boxes indicate missed and false detections respectively.
    }
    
\vspace{5mm}

    \label{fig:color_thermal}
\end{figure}

Figure \ref{fig:color_thermal} compares the two unimodal outputs with multimodal model. The RGB image (left) fails to detect the pedestrian in the dark area marked by a green box in the first row. Both the thermal (middle) and multimodal (right) model can detect it successfully. Two pedestrians are missed in thermal image output in the second row but correctly detected in RGB and multimodal outputs. RGB image output produces a false positive marked by a red box in the third row, but it is not present in thermal and fusion outputs. These three scenarios illustrate that fusion can effectively combine both RGB and thermal inputs.

\subsection{Confidence Estimation}
\new{The problem of pedestrian detection has safety critical implications in the field of autonomous driving. Thus, it is necessary to have a confidence estimation for each prediction. The confidence of a prediction may be low due to various reasons including an out of distribution sample, occlusion or poor visibility of the pedestrian and similarity of appearance with negative samples such as poles. Downstream modules such as fusion, tracking or motion planning can aggregate the confidence values from various object detections to perform better system level predictions and assess the risk of the situation. Thus we aim to implement a confidence estimation in our model.}

\new{
Typically, object detection networks employ a softmax whose output can be interpreted as posterior probability of classification. Class corresponding to the maximum probability is chosen for the prediction and sometimes the value of this maximum probability is used as confidence. However, softmax saturates quickly due to the exponential function and tends to be overconfident in practice with poor performance in the case of adversarial attacks and out of distribution samples \cite{nguyen2015deep, lakshminarayanan2017simple}. Bayesian Neural Networks can be approximated by inference time drop outs \cite{gal2016dropout} providing a rigorous uncertainty within Bayesian formalism. Ensemble models \cite{lakshminarayanan2017simple} is another popular method to estimate the confidence value. However, these methods significantly increase the complexity of the model and thus we chose a simple recent method ConfidNet based on auxiliary confidence estimation model \cite{corbiere2021confidence}. It can be easily implemented by addition of a small confidence estimation decoder. The inference time increases incrementally as the encoder computation is reused.}

\new{We adapt the ConfidNet auxiliary decoder to make use of the fused feature maps output from the SCoFA module. We train the auxiliary decoder to predict the true class probability of our model in the CVC-14 dataset as per the ConfidNet training protocol. The model prediction accuracy is not affected by the addition of the confidence module. The confidence value may be useful to improve the detection performance by performing an additional thresholding but our intention is to use it for downstream modules. Thus, we perform an analysis of the statistics of predictions at different confidence levels. The confidence value is normalized between 0 and 1, higher confidence value is expected to produce a better prediction. }

\begin{figure}
    \captionsetup{singlelinecheck=false, font=small, belowskip=4pt}
    \centering
    \includegraphics[width=0.46\textwidth]{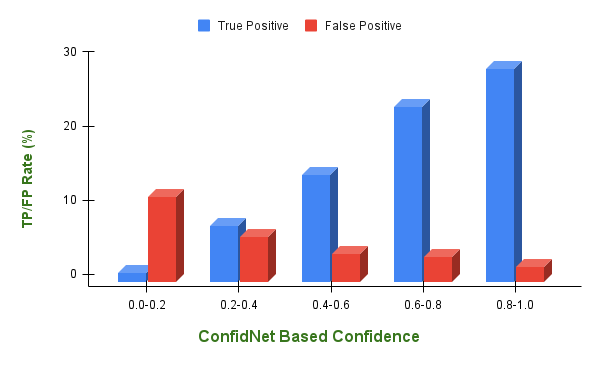} 
    \caption{\new{Confidence statistics of our model on CVC-14  dataset \cite{hwang2015multispectral}.}}
    \label{Uncertainty_Histogram}
\end{figure}

\new{
We split the confidence range into five bins of step size 0.2 and plot the true positive and false positive rate of our predictions in Figure \ref{Uncertainty_Histogram}. We first observe that the confidence value is working well at the extremes. At high confidence range of 0.8 to 1, the predictions are nearly 15 times more likely to be a true positive than a false positive. At low confidence range, the predictions are nearly 10 times more likely to be a false positive than a true positive. The range of 0.2 to 0.4 is the most ambiguous where a true positive and false positive are equally likely. After this range, true positive rate rapidly increases and false positive rate rapidly diminishes.
}

\section{Conclusion} 
\label{sec:conc}

In this work, we have developed a novel end-to-end deep learning architecture for multimodal pedestrian detection. 
We have introduced the \textit{Multimodal Feature Embedding Module} for the embedding of the two uni-modal encoder features into a lower-dimensional multimodal feature vector. It makes use of attention-guided graph networks to extract features representing the structural properties of pedestrians.
We also introduced the \textit{Spatio-Contextual Feature Aggregation} Module, which combines the spatial information obtained from its channel-wise attention block and the contextual information obtained from its $4$Dir-IRNN block. We obtain state-of-the-art results on three public multimodal benchmarks, namely KAIST, CVC-$14$, and UTokyo datasets.
\new{In future work, we plan to develop a single model which can be trained and tested across all the three datasets addressing the differences in resolutions and fields of view. We also plan to exploit the temporal information by feeding in multiple frames in the video sequence.}

\bibliographystyle{IEEEtran}
\bibliography{references/egbib}

\vspace{5mm}

\begin{IEEEbiography}
[{\includegraphics[width=1in,height=1.5in,clip,keepaspectratio]{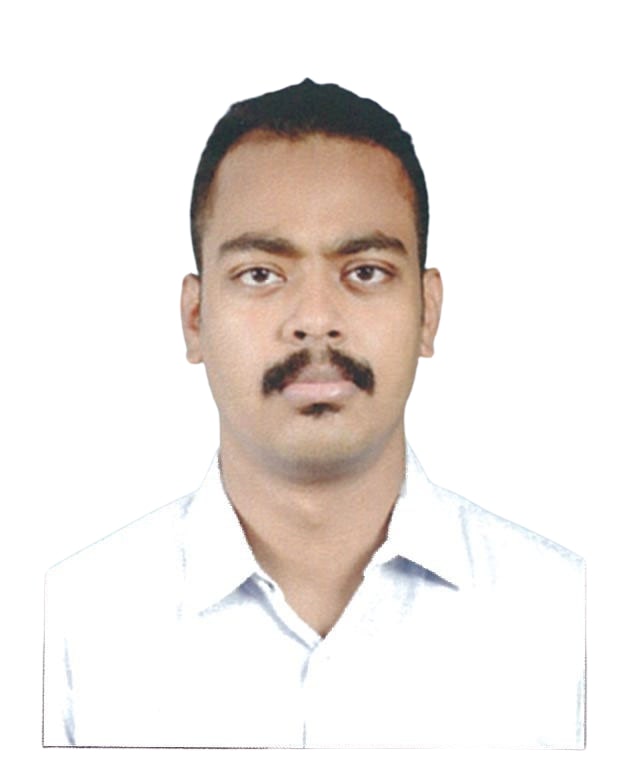}}]{Kinjal Dasgupta} received his B.Tech. degree in Electrical Engineering from the West Bengal University of Technology, India, in 2020. He is currently holding a Research Position at Indian Statistical Institute, Kolkata, India. His areas of current research interest are Scene Understanding and the relevant problems of Computer Vision related to Supervised/Unsupervised Learning, Domain Adaptation and Transfer Learning.
\end{IEEEbiography}

\begin{IEEEbiography}
[{\includegraphics[width=1in,height=1.5in,clip,keepaspectratio]{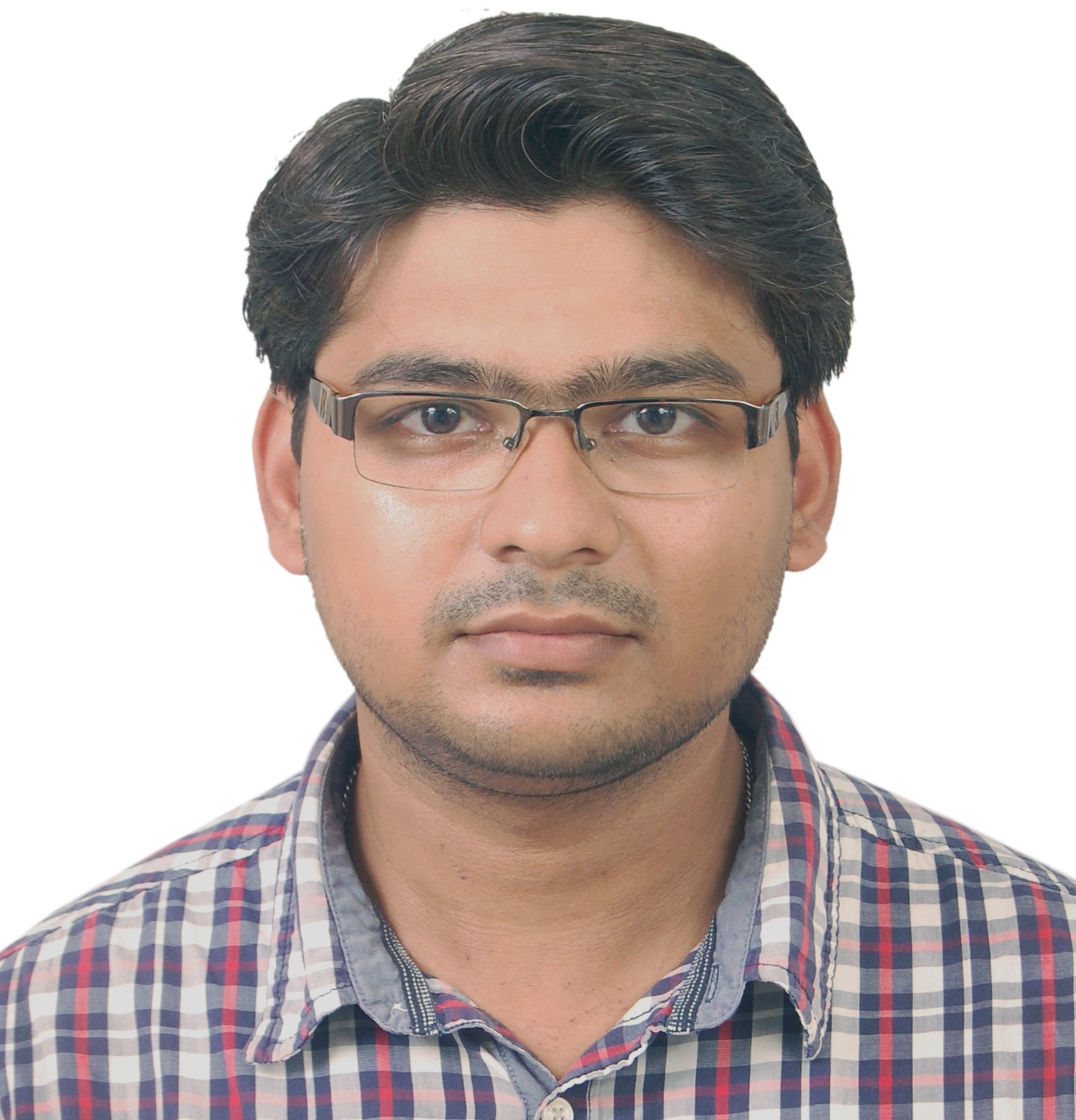}}]{Arindam Das} is a Lead Engineer in the department of Detection Vision Systems (DVS) at Valeo India. He is elected as Valeo Expert for 2022 in AI. He is currently involved in deep learning based algorithm development activities for autonomous driving systems. He has more than $8$ years of industry experience in computer vision, deep learning \& document analysis. He has authored $13$ peer reviewed publications \& $34$ patents. His current area of research interest includes self-supervised learning, domain adaptation, image restoration \& multimodal learning.
\end{IEEEbiography}

\begin{IEEEbiography}[{\includegraphics[width=1in,height=1.25in,clip,keepaspectratio]{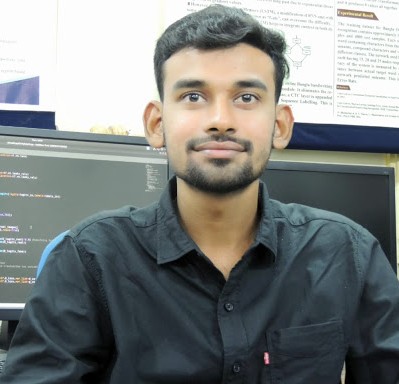}}]{Sudip Das} joined the Indian Statistical Institute in a research position of its Computer Vision and Pattern Recognition Unit at Kolkata after obtaining his B. Tech degree in Computer Science and Engineering from the West Bengal University of Technology (WBUT), India, in 2017. His principal area of research focuses on the relevant problems of Unsupervised Leaning, Curriculum Learning, Transfer Learning and Domain Adaptation. 
He is also passionate to work on the various problems of autonomous driving. In particular, he did some research in Computer Vision, with the goal of Detecting, Segmenting and Pose Estimating of objects in Images or videos.
\end{IEEEbiography}


\begin{IEEEbiography}
[{\includegraphics[width=1in,height=1.25in,clip,keepaspectratio]{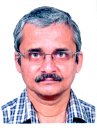}}]{Ujjwal Bhattacharya} is an Associate Professor of  Computer Vision and Pattern Recognition Unit (CVPRU) of Indian Statistical Institute (ISI), Kolkata. He is currently serving as the Head of the Computer and Statistical Service Centre (CSSC) of ISI. He is a senior member of the IEEE, a life member of IUPRAI, Indian unit of the IAPR and an Associate Member of the Centre for Artificial Intelligence and Machine Learning (CAIML) of ISI. He worked as a Co-Guest Editor of a few Special Issues of International Journals. He is a regular reviewer of various reputed International Journals and Conferences. His current research interests include Machine Learning, Computer Vision, Image Processing, Document Processing, Handwriting Recognition etc.
\end{IEEEbiography}

\begin{IEEEbiography}
[{\includegraphics[width=1in,height=1.25in,clip,keepaspectratio]{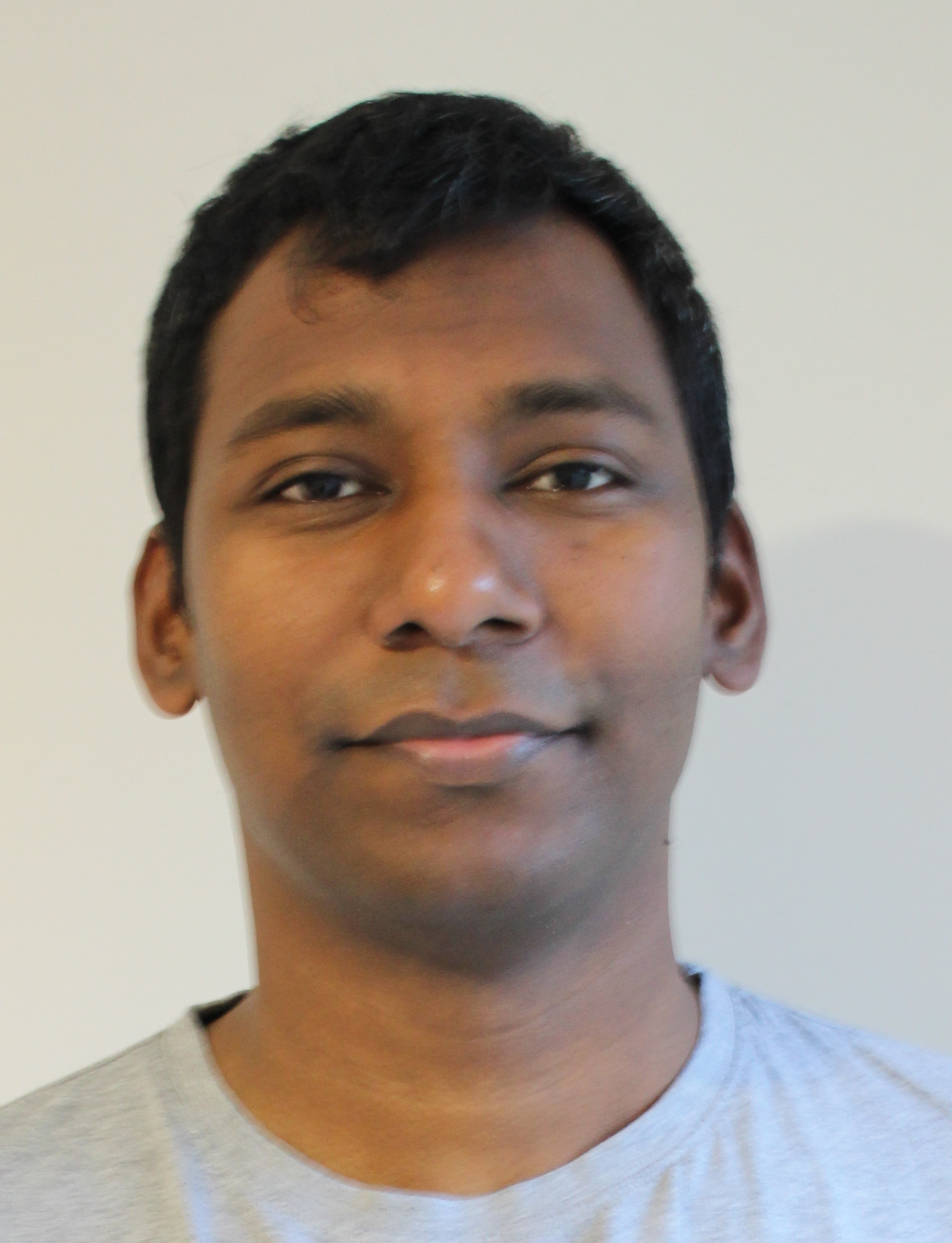}}]{Senthil Yogamani}
is an Artificial Intelligence architect and holds a director level technical leader position at Valeo Ireland. He leads the research and design of AI algorithms for various modules of autonomous driving systems. He has over 15 years of experience in computer vision and machine learning including 13 years of experience in industrial automotive systems. He is an author of 100+ publications with 2700 citations and 100+ inventions with 70 filed patent families. He serves on the editorial board of various leading IEEE automotive conferences including ITSC and IV and advisory board of various industry consortia including Khronos, Cognitive Vehicles and IS Auto. He is a recipient of the best associate editor award at ITSC 2015 and best paper award at ITST 2012.
\end{IEEEbiography}

\end{document}